\documentclass{article}

\usepackage{bbm}
\usepackage{bm}
\usepackage{amsmath}
\usepackage{amsbsy}
\usepackage{amssymb}
\usepackage{amsfonts}
\usepackage[ruled, vlined, linesnumbered]{algorithm2e}
\usepackage{graphicx, subfigure}
\usepackage{multirow}
\usepackage{hhline}
\usepackage{color}
\usepackage{booktabs}
\usepackage{tikz}
\usepackage{pgfplots}

\usepackage{comment}
\usepackage{rotating}

\newtheorem{theorem}{Theorem}

\newtheorem{corollary}{Corollary}
\newtheorem{lemma}{Lemma}
\newtheorem{proposition}{Proposition}

\newcommand{\qed}{\hfill $\square$}
\newcommand{\proof}{\noindent \emph{Proof. }}

\renewcommand{\emptyset}{\varnothing}

\newcommand{\Sf}[1]{\mathsf{#1}}

\newcommand{\ZZ}{\mathbbm{Z}}

\renewcommand{\gg}{\Sf{g}}

\newcommand{\kk}{\kappa_B}
\newcommand{\kd}{\kappa_{\Delta}}
\newcommand{\kdp}{\kappa_{\Delta_p}}

\newcommand{\kdc}{\kappa_{\Delta_{|c^*|}}}

\newcommand{\qa}{\textsc{QuAcq}\xspace}

\newcommand{\fscope}{\texttt{FindScope}\xspace}
\newcommand{\fc}{\texttt{FindC}\xspace}
\newcommand{\genex}{\texttt{Generate\-Example}\xspace}
\newcommand{\genexb}{\genex.\texttt{basic}\xspace}
\newcommand{\genexc}{\genex.\texttt{cutoff}\xspace}

\newcommand{\Time}{\texttt{time}\xspace}
\newcommand{\cutoff}{\mathsf{cutoff}\xspace}

\newcommand{\modelseeker}{\emph{ModelSeeker}\xspace}
\newcommand{\arnold}{\emph{Arnold}\xspace}

\newcommand{\yes}{\textit{yes}\xspace}
\newcommand{\no}{\textit{no}\xspace}
\newcommand{\true}{\textit{true}\xspace}
\newcommand{\false}{\textit{false}\xspace}

\newcommand{\qao}{\textsc{QuAcq1}\xspace}
\newcommand{\qat}{\textsc{QuAcq2}\xspace}
\newcommand{\qatb}{\qat.\texttt{basic}\xspace}
\newcommand{\qatc}{\qat.\texttt{cutoff}\xspace}

\SetKw{myproc}{procedure}
\SetKw{myfunc}{function}
\SetKw{Inp}{in}
\SetKw{Outp}{out}
\SetKw{InOutp}{in out}
\SetKw{F}{false}
\SetKw{T}{true}
\SetKw{lAnd}{and}
\SetKw{Exit}{exit}

\title{Partial Queries for Constraint Acquisition\thanks{This paper 
		extends and corrects the work published in \cite{BessiereCHKLNQW13}. }
	\thanks{This  work has been  funded by the  
		European Community 
		project FP7-284715 ICON.}
}
\author{
	Christian Bessiere\\CNRS, University of Montpellier\\ France\and 
	Clément Carbonnel\\CNRS, University of Montpellier\\ France\and 
	Anton Dries\\DTAI, KU Leuven\\ Belgium\and 
	Emmanuel Hebrard\\LAAS-CNRS\\ Toulouse,  France\and 
	George Katsirelos\\INRA Toulouse\\ France\and
	Nadjib Lazaar\\University of Montpellier\\ France\and 
	Nina Narodytska\\Samsung Research America\\ USA\and
	Claude-Guy Quimper\\University Laval\\ Quebec City,  Canada\and
	Kostas Stergiou\\University of Western Macedonia\\ Kozani, Greece \and
	Dimosthenis C. Tsouros\\University of Western Macedonia\\ Kozani, Greece \and
	Toby Walsh\\NICTA, UNSW\\ Sydney, Australia 
}

\begin{document}

\maketitle

\begin{abstract}
	Learning constraint networks is known to require a 
	number of membership queries  exponential in the number of variables. 
	In this paper, we learn constraint networks by asking the user
	partial queries. That is, we ask the user to classify assignments to
	subsets of the variables as positive or negative. We provide an
	algorithm, called \qa, that, given a negative example, focuses onto a constraint of the 
	target network in a number of queries logarithmic in
	the size of the example. The whole constraint network can then be learned 
	with a polynomial number of partial queries. 
	We  give  information
	theoretic  lower  bounds  for  learning  some  simple
	classes  of  constraint  networks  and  show  that  our
	generic algorithm is optimal in some cases. Finally
	we evaluate our algorithm on some benchmarks.
\end{abstract}

\section{Introduction}\label{sec:introduction}

Constraint programming (CP) has been  more and more  used 
to solve combinatorial problems in industrial applications. 
One of the strengths of CP is that it is declarative, which 
means that the user specifies the problem 
as a CP model, and the solver finds solutions. 
However, it appears that specifying the CP model 
is not that easy for non-specialists. 
Hence, the modeling phase constitutes 
a major bottleneck in the use of CP. 
Several techniques have been proposed to tackle this
bottleneck. 
In \emph{Conacq.1}
\cite{besetalCP04modelling,besetalECML05,BessiereKLO17}, 
the user provides examples of
solutions (positive) and non-solutions (negative). 
Based on these positive and negative examples, 
the system learns a set of
constraints that correctly classifies all examples given so
far. This is a form of \emph{passive} learning. 
%
%
A  passive learner based on inductive logic programming is presented in
\cite{DBLP:conf/ictai/LallouetLMV10}. This system requires  background knowledge on
the structure of the problem  to learn a representation of the
problem correctly classifying the examples. 
In  \modelseeker \cite{DBLP:conf/cp/BeldiceanuS12}, the user provides positive examples to the
system, which arranges each of them as a  matrix and 
identifies constraints in the global
constraints catalog (\cite{beletalTR05}) 
that are satisfied by particular subsets of variables in all 
the examples. 
Such particular subsets are for instance rows or columns. 
The candidate constraints are ranked and proposed to the user for selection. 
This ranking/selection   combined with the
representation of examples as matrices allows \modelseeker to 
quickly find a good model when the problem has an underlying 
matrix structure. 
More recently, a passive learner called \arnold has been proposed \cite{KumarTR19}. 
\arnold takes positive examples as input and  returns an integer 
program that accepts these examples as solutions. 
\arnold relies on a tensor-based language for describing polynomial 
constraints between multi-dimensional vectors. As in \modelseeker, 
the problem needs to have an underlying  matrix structure. 
\emph{Conacq.1} is thus the only passive learner that can learn constraints 
when the problem does not have a specific structure. 

%

By contrast, in an \emph{active learner} like  \emph{Conacq.2}
\cite{besetalIJCAI07queries,BessiereKLO17}, 
the system proposes examples to the user
to  classify as solutions or non solutions. 
Such questions are called \emph{membership queries} \cite{angluin87}. 
In applications where we need a proof that the learning system has converged to 
the target set of constraints, active learning is a good candidate because it 
can significantly decrease the number of examples necessary to
converge. 
For instance, a few years ago, the Normind company hired a constraint
programming specialist to transform their 
expert system for detecting failures in electric
circuits in Airbus airplanes into a constraint model in
order to make it more efficient and easier to maintain. 
An active learner can do this by automatically interacting with the expert system. 
As another example, 
active learning was used to  build a constraint model
that encodes non-atomic actions of a robot (e.g., catch a ball) by
asking queries of the simulator of the robot  \cite{pauetalICTAI08}. 
Such active learning introduces two computational challenges. 
First, how does the system generate a useful query? 
Second, how many queries are needed for 
the system to converge to the target set of constraints? 
It has been 
shown that 
the number of membership queries required to converge to the target set of
constraints can be exponentially large 
\cite{beskor12,BessiereKLO17}. 

In this paper, we propose \qa (for Quick Acquisition), an
active learner that asks
the user to classify {\em partial} queries as positive or negative. 
Given a negative example, \qa is able to
learn a constraint of the target constraint network in a 
number of queries logarithmic in the number of variables. 
As a result,  \qa   converges on
the target constraint network in a polynomial number of queries. 
In fact, we identify information theoretic
lower bounds on the complexity of learning constraint networks 
that show that \qa is optimal on some simple  
languages and close to optimal 
on  others.
One application for \qa would be
to learn a general purpose model. In constraint programming,
a distinction is made between model and data.
For example, in a sudoku puzzle, the model
contains generic constraints like each subsquare
contains a permutation of the numbers. 
The data, on the other hand, gives the
pre-filled squares for a specific puzzle. 
As a second example, in a time-tabling problem, the
model specifies generic constraints like no
teacher can teach multiple classes at the same
time. The data, on the other hand, 
specifies particular room sizes, and teacher availability
for a particular time-tabling problem instance. 
The cost of learning the model can then be amortized over the
lifetime of the model. 
\qa has several advantage. 
First, it is the only approach ensuring the property of  convergence in a polynomial 
number of queries. 
Second,  as opposed to existing
techniques, the user does not need to give positive
examples. This might be useful 
if the problem has not yet been solved, so 
there are no examples of past solutions.  
Third,  \qa   learns any kind of network, 
whatever the constraints are organized in  a specific structure or not. 
Fourth,  \qa can be used when part of the network is already known from 
the user or from another learning technique. Experiments  show 
that the larger the amount of known constraints, the fewer the queries required to 
converge on the target network.

The rest of the paper is organized as follows. 
Section \ref{sec:background} gives the necessary definitions to 
understand the technical presentation. 
Section \ref{sec:related-qa} describes the differences between the algorithm 
presented in this paper and the version in \cite{BessiereCHKLNQW13}. 
Section \ref{sec:qa} presents \qa, the algorithm that 
learns constraint networks by asking partial queries. 
In Section \ref{sec:formal}, we show that \qa behaves optimally on 
some simple languages. 
Section \ref{sec:exp}
presents an experimental evaluation of  \qa. 
Section \ref{sec:conclusion} concludes the paper. 

\section{Background}
\label{sec:background}

\newcommand{\C}{\Sf{c}}
\newcommand{\B}{\Sf{b}}
\newcommand{\CC}{\Sf{C}}
\newcommand{\LL}{\Sf{L}}

\newcommand{\Q}{\mathcal{Q}}
\newcommand{\X}{\mathcal{X}}
\newcommand{\D}{\mathcal{D}}
\newcommand{\K}{\Sf{K}}
\newcommand{\I}{I}
\newcommand{\R}{R}
\newcommand{\xx}{x}
\newcommand{\gen}{\psline{->}}
\newcommand{\cont}{\uput{0}[90]}
\newcommand{\point}{\psline[linestyle=dashed]{->}}
\newcommand{\pointeur}{\pscurve[linestyle=dashed]{->}}
\newcommand{\genf}{\psline[linewidth=0.01]{->}}
\newcommand{\emphdef}[1]{\textbf{#1}}
\newcommand{\genl}{\texttt{Generalise}\xspace}
\newcommand{\alldiff}{\texttt{AllDiff}\xspace}
\newcommand{\finc}{\texttt{FindCouple}\xspace}

The learner and the user need to share some  common knowledge to
communicate. We suppose this common knowledge,
called the \emph{vocabulary}, is a (finite) set of $n$
variables $X$ and a  domain $D=\{D(X_1), \ldots, D(X_n)\}$, where
$D(X_i)\subset\ZZ$ is
the finite set of values for $X_i$.  
A \emph{constraint}  $c$ is defined by a sequence of variables $scp(c)\subseteq X$, 
called {\em the constraint scope}, and a relation $rel(c)$ over $\ZZ$ specifying 
which sequences of $|scp(c)|$ values are allowed for the variables $scp(c)$. 
We will use the notation $var(c)$ to refer to the \emph{set} of  variables 
in  $scp(c)$, and we abusively call it 'scope' too when no confusion is possible.   
%
%
A \emph{constraint network} (or simply \emph{network}) is a set $C$ of
constraints on the vocabulary $(X,D)$. 
An assignment $e_Y \in D^Y$, 
where  $D^Y=\Pi_{X_i \in Y} D(X_i)$,  
is called a partial assignment when $Y \subset X$ and 
a complete assignment when $Y = X$. 
An assignment $e_Y$ on a set of variables $Y\subseteq X$
\emph{is rejected by} a  constraint $c$ 
(or $e_Y$ \emph{violates}  $c$) if $var(c)\subseteq Y$ and the
projection $e_Y[scp(c)]$ of $e_Y$ on the variables $scp(c)$ is not in $rel(c)$. 
If $e_Y$ does not violate $c$, it \emph{satisfies} it.  
An assignment $e_Y$ on $Y$ \emph{is accepted by} $C$ 
if and only if it does not violate any constraint in $C$. 
An assignment on $X$ that is accepted by $C$ is a \emph{solution} of $C$. 
We write $sol(C)$ for the set of solutions of $C$.  
We write $C[Y]$ for the set of constraints in $C$ whose scope is included in $Y$, 
and $C_Y$ for the set of constraints in $C$ whose scope is exactly $Y$. 
We say that two networks $C$ and $C'$ are \emph{equivalent} if $sol(C)=sol(C')$.

In addition to the vocabulary, the learner owns  
a \emph{language} $\Gamma$ of bounded arity 
relations from which it can build constraints 
on specified sets of variables. 
Adapting terms from machine learning, 
the \emph{constraint basis}, denoted by $B$, is a set of  constraints built
from the constraint language $\Gamma$ on the  vocabulary $(X,D)$
from which the learner builds a constraint network.

The  {\em target network} is a network  $T$ such that 
$T\subseteq B$ and for any 
example $e\in D^{X}$, $e$ is a solution of $T$ if and only 
if $e$ is a  solution of the problem that the user has in mind. 
A  \emph{membership query} $ASK(e)$ takes as input a \emph{complete}  
assignment $e$ in $D^X$ and asks the user to classify it.   
The answer to $ASK(e)$ is \yes  if and only if
$e \in sol(T)$.
A {\em partial query} $ASK(e_Y)$, 
with $Y\subseteq X$,  takes as input a \emph{partial}  
assignment $e$ in $D^Y$ and asks the user to classify it. 
The answer to $ASK(e_Y)$ is \yes if and only if $e_Y$ 
does not violate any constraint in $T$. 
It is important to observe that "$ASK(e_Y)$=\yes" does not 
mean that $e_Y$ extends to a solution of $T$, which would put 
an NP-complete problem on the shoulders of the user. 
A classified assignment $e_Y$ is called positive or negative \emph{example}
depending on whether $ASK(e_Y)$ is \yes or \no.  
For any assignment  $e_Y$ on $Y$, $\kk(e_Y)$ denotes the set of all
constraints in $B$  that reject $e_Y$. We will also use $\kd(e_Y)$ to  
denote the set of constraints in a given set $\Delta$ that reject $e_Y$.

We now define \emph{convergence}, which is the  constraint acquisition
problem we are interested in. 
Given a set $E$ of (partial) examples labeled by
the user \yes or \no, 
we say that a  network $C$ agrees with $E$ if  $C$
accepts all examples labeled \yes in  $E$ and does not accept those
labeled \no. 
The learning process has \emph{converged} on the network 
$L\subseteq B$ if $L$ agrees with $E$ and for every other network
$L'\subseteq B$ agreeing with  $E$, we have  $sol(L')= sol(L)$. 
We are thus guaranteed that $L$ is equivalent to $T$. 
It is important to note that $L$ is not necessarily unique and equal to $T$. 
This is because of   \emph{redundant} constraints. 
Given a set $C$ of constraints, a constraint $c\notin C$ is redundant 
wrt  $C$ if $sol(C)=sol(C\cup\{c\})$.  
If a constraint $c$  from $B$ is redundant wrt $T$, the network 
$T\cup \{c\}$ is equivalent to~$T$.  

In the algorithms presented in the rest of the paper we will use 
the \emph{join} operation, denoted by $\Join$. 
Given two sets of constraints $S$ and $S'$, the join of 
$S$ with $S'$ is the set of non-empty constraints obtained by 
pairwise conjunction of a constraint in $S$ with a constraint in $S'$. 
That is, 
$S\Join S'=\{c\land c'\mid c\in S, c'\in S', c\land c'\not\models\bot\}$. 
A constraint belonging to the basis $B$ will be called 
\emph{elementary}  in contrast to a constraint composed 
of the conjunction of several elementary constraints, which will 
be called \emph{conjunction}. 
A conjunction will also sometimes be referred to as a \emph{set} of 
elementary constraints. 
Given a set $S$ of conjunctions, we will use the notation 
$S_p$ to refer to the subset of $S$ containing only 
the conjunctions composed 
of at most $p$ elementary constraints. 
Finally, a \emph{normalized} network is a network 
that does not contain conjunctions of constraints on any scope, 
that is, all its constraints are elementary.

\section{\qat versus \qao}\label{sec:related-qa}

A first version of \qa was published in \cite{BessiereCHKLNQW13}. 
From now on let us call it \qao. 
That version was devoted to normalized constraint networks, that is, networks 
for which there does not exist any pair of constraints with scopes included 
one in the other. 
In addition, \qao was not taking as assumption that the target network 
is a subset of constraints from the basis. 
As a consequence, 
when the target network was not a subset of the basis, \qao was 
either learning a wrong network or was subject to a "collapse" state. 
When the target network was a subset of the basis, \qao was 
asking redundant (i.e., useless) queries. 
In \qat, the  problem of  constraint acquisition is formulated 
in a way that is more in line with standard concept learning 
\cite{AngluinFP92,RaedtPT18}. 
The target network is a subset of the constraints in the basis. 
As a consequence, an active learner such as \qat will always return 
the last possible constraint network given a set of examples already classified. 
It will never collapse. 
The second difference with \qao is that \qat does not require that the 
target network is normalized. 
\qat can learn any type of constraint network.

\section{Constraint Acquisition with Partial Queries}
\label{sec:qa}

We propose \qat, a novel active learning algorithm. \qat
takes as input a basis $B$ on a vocabulary $(X,D)$. It 
asks partial queries of the user 
until it has converged 
on a constraint network
$L$ equivalent to the target network $T$. 
When a query is answered \yes, constraints
rejecting it are removed from $B$.  When a query is answered \no, \qat
enters a loop (functions \fscope and \fc) that will end by the
addition of a constraint to $L$.

\subsection{Description of \qat}

\qat (see Algorithm \ref{alg:qa}) initializes the network $L$ it will learn
to the empty set (line \ref{qa:init}). 
In line \ref{qa:gen-sol}, \qat calls function \genex that computes an 
assignment $e_Y$ on a subset of variables $Y$ 
satisfying the constraints of $L$ 
that have a scope included in $Y$, 
but violating at least one constraint from $B$.\footnote{For 
	this task, the constraint solver needs to be able 
	to express the negation of the constraints in $B$. This is not a problem 
	as we have only bounded arity constraints in $B$.}
We will see later that there are multiple ways to design function \genex. 
If there does not exist any pair $(Y,e_Y)$ accepted by $L$ and 
rejected by $B$  (i.e., \genex returns $\bot$), then all
constraints in $B$ are implied by  $L$, and we have converged   (line \ref{qa:converge}).  
If we have
not converged, we propose the example $e_Y$ to the user, who will answer
by \yes or \no (line \ref{qa:ask}).  
If the answer is \yes, we can remove from $B$ the set $\kk(e_Y)$ of
all constraints in $B$ that reject $e_Y$ (line~\ref{qa:ask}). 
If the answer is \no, we are sure that $e_Y$ violates at least one
constraint of the target network $T$. 
We then call the function \fscope to
discover the scope $S$ of one of these violated
constraints, and the procedure  
\fc will learn (that is, put in $L$) 
at least one constraint of $T$ whose scope is in $S$
(line \ref{qa:findc}). 

\SetKwInOut{Input}{In}
\SetKwInOut{Output}{Out}
\SetKwInOut{InOutput}{In Out}

\begin{algorithm}[tb]
	\Input{A basis $B$}
	\Output{A learned network $L$}
	\Begin{
		$L \gets \emptyset$\label{qa:init}\;
		\While{true\label{qa:loop}}{
			$e_Y\gets \genex(X,L,B)$\label{qa:gen-sol}\;    
			\lIf{$e_Y = \bot$}{
				\Return ``convergence on $L$''\label{qa:converge}} 
			\uIf{$ASK(e_Y) = \yes$\label{qa:ask} }{ 
				$B\gets B\setminus \kappa_B(e_Y)$\label{qa:remove}\;
			}
			\lElse{
				$ \fc(e_Y,\fscope(e_Y,\emptyset,Y),L)$\label{qa:findc}}
		}
	}
	\caption{\qat 
	}
	\label{alg:qa}
\end{algorithm}

The recursive function \fscope (see Algorithm \ref{alg:fscope}) 
takes as parameters an example $e$ and 
two sets $R$ and $Y$ of variables. 
An invariant of \fscope is that $e$ violates at least one
constraint whose scope is a subset of $R\cup Y$. 
A second invariant is that \fscope always returns 
a subset of $Y$ that is also 
the subset of the scope of a constraint violated by $e$. 
If  there is at least one  constraint in $B$ 
rejecting $e[R]$ (i.e., $\kk{(e[R])}\neq \emptyset$, line \ref{qa:elu:doweask}), we ask 
the user whether $e[R]$ is positive
or not (line \ref{qa:elu:ask}). 
If the answer is \yes, we can remove all the constraints that reject
$e[R]$ from $B$. 
If the answer is \no, we are sure that $R$ itself contains 
the scope of a constraint of  $T$ rejecting $e$. 
As $Y$ is not needed to cover that scope, we return the empty set (line  \ref{qa:elu:emptyset}). 
We reach line \ref{qa:elu:singleton} only in case $e[R]$ does not
violate any constraint. We know that
$e[{R\cup Y}]$ violates a constraint. Hence, if $Y$ is a singleton,
the variable it contains necessarily belongs to the scope of a
constraint that violates $e[{R\cup Y}]$. The function returns $Y$. 
If none of the return conditions are satisfied, the set $Y$ is
split in two balanced parts $Y_1$ and $Y_2$ (line \ref{qa:elu:split}) and we apply a
technique similar
to \textsc{QuickXplain} (\cite{junkerAAAI04}) to elucidate the
variables of a constraint violating $e[{R\cup Y}]$ in
a logarithmic number of steps (lines \ref{qa:elu:call1} and \ref{qa:elu:call2}). 
In the first recursive call, if $R\cup Y_1$ does not contain any scope 
$S$ of constraint rejecting $e$, 
\fscope returns a subset $S_1$ of such a scope such that $S_1=S\cap Y_2$ and 
$S\subseteq R\cup Y$.  
In the second recursive call, the variables returned in $S_1$ are 
added to  $R$. 
if $R\cup S_1$ does not contain any scope 
$S$ of constraint rejecting $e$, 
\fscope returns a subset $S_2$ of such a scope such that $S_2=S\cap Y_1$ and 
$S\subseteq R\cup Y$.  
The rationale of  lines \ref{qa:elu:kappa1} and \ref{qa:elu:kappa2} 
is to avoid entering a recursive call to \fscope when 
we know the answer to the query in line \ref{qa:elu:ask} of that call 
will necessarily  be \no. 
It happens when all  the constraints 
rejecting $e[R\cup Y]$ have a scope  included in the set of variables 
that will be $R$ inside that call (that is, 
$R\cup Y_1$ for the call in line \ref{qa:elu:call1}, 
and $R$ union the output of line \ref{qa:elu:call1} for the call 
in line \ref{qa:elu:call2}). 
Finally, line \ref{qa:elu:end} of \fscope returns the union of the two 
subsets of variables returned by the two recursive calls, as we know 
they all belong to the same scope of a constraint of $T$ rejecting $e$.

\begin{algorithm}[tb]
	\Input{An example $e$ ; Two scopes $R,Y$}
	\Output{The scope of a constraint in $T$}
	\Begin{
		\If{$\kappa_B(e[R])\neq\emptyset$\label{qa:elu:doweask} }{
			\lIf{$ASK(e[{R}])=\yes$\label{qa:elu:ask}}{
				$B\gets B\setminus \kappa_B(e[R])$\label{qa:elu:cleanB}}
			\lElse{\Return $\emptyset$\label{qa:elu:emptyset}}
		}
		\lIf{$|Y|=1$}{\Return Y\label{qa:elu:singleton}}
		split $Y$ into $<Y_1,Y_2>$ such that $|Y_1|=\lceil
		|Y|/2\rceil$
		\label{qa:elu:split}\; 
		\lIf{$\kappa_B(e[R\cup Y_1])=\kappa_B(e[R\cup Y])$\label{qa:elu:kappa1}}{$S_1\gets \emptyset$}
		\lElse{$S_1 \gets \fscope(e, R\cup Y_1, Y_2)$\label{qa:elu:call1}}
		\lIf{$\kappa_B(e[R\cup S_1])=\kappa_B(e[R\cup Y])$\label{qa:elu:kappa2}}{$S_2\gets \emptyset$}
		\lElse{$S_2 \gets \fscope(e,R\cup S_1, Y_1)$\label{qa:elu:call2}}
		$\Return\ S_1\cup S_2$\label{qa:elu:end}\;
	}
	\caption{Function \fscope}
	\label{alg:fscope}
\end{algorithm}

The function \fc (see Algorithm \ref{alg:detect}) takes 
as parameter $e$ and $Y$, 
$e$ being the negative example that led 
\fscope to find that there is a
constraint from the target network $T$ over the scope $Y$. 
The set $\Delta$ is initialized to all candidate constraints, 
that is, the set $B_Y$ of all constraints from $B$ 
with scope exactly $Y$ (line \ref{qa:det:init}).
As we know from \fscope that 
there will be a constraint with scope $Y$ in $T$, 
we join $\Delta$ with the set of constraints of scope  $Y$ 
rejecting $e$ (line \ref{qa:det:join}). 
In line \ref{qa:det:select-ex}, an example $e'$ is chosen in such a way
that $\Delta$  contains both constraints satisfied by $e'$ and constraints
violated by $e'$. 
If no such example exists (line \ref{qa:det:nil}), this means that
all constraints in $\Delta$ are equivalent wrt $L[Y]$. Any of them
is added to $L$ and $B$ is emptied of all its constraints with 
scope $Y$ (line \ref{qa:det:return}).   
If a suitable example $e'$ was found, it is proposed to the user for
classification (line \ref{qa:det:ask}). If $e'$ is classified positive, 
all constraints rejecting it are removed from  $\Delta$ and $B$ (line
\ref{qa:det:cleanBD}). 
Otherwise we call \fscope to seek constraints 
with scope strictly included in $Y$ that violate $e'$ (line \ref{qa:det:findsc}). 
If \fscope returns the scope of such a constraint, 
we recursively call  \fc 
to find that smaller arity constraint before the one having scope $Y$ 
(line \ref{qa:det:findc}). 
If \fscope has not found such a scope (that is, it returned $Y$ itself), 
we do the same join as in line \ref{qa:det:join} to 
keep in  $\Delta$ only constraints
rejecting the example $e'$ (line~\ref{qa:det:join2}). 
Then,  we continue the loop of line \ref{qa:det:loop}.

\begin{algorithm}[tb]
	\Input{An example $e$ ; A scope $Y$}
	\InOutput{The network $L$}
	\Begin{
		$\Delta\gets 
		B_Y$\label{qa:det:init}\;
		$\Delta\gets \Delta\Join\kd(e)$\label{qa:det:join}\;
		\While{true\label{qa:det:loop}
		}{
			choose $e'_Y$ in $sol(L[Y])$ such that $\emptyset \subsetneq \kd(e'_Y) \subsetneq\Delta$\label{qa:det:select-ex}\; 
			\uIf{$e'_Y=\bot$\label{qa:det:nil}}{  
				pick $c$ in $\Delta$\;\label{qa:det:pick}
				$L\gets L\cup\{c\}; B\gets B\setminus B_Y$; \Exit;\label{qa:det:return}}
			\Else{
				\uIf{$ASK(e'_Y) = \yes$\label{qa:det:ask}}{
					$\Delta\gets \Delta\setminus \kd(e'_Y); B\gets B\setminus \kk(e'_Y)$;\label{qa:det:cleanBD}
				}\Else{
					$S\gets \fscope(e'_Y,\emptyset,Y)$\label{qa:det:findsc}\;
					\lIf{$S\subsetneq Y$}{$\fc(e'_Y,S,L)$\label{qa:det:findc}}
					\lElse{
						$\Delta \gets \Delta\Join \kd(e'_Y)$\label{qa:det:join2}}}
		}}
	}
	\caption{Procedure \fc }
	\label{alg:detect}
\end{algorithm}

At  this point we can make an observation on the kind of response the user is 
able to give. \qat is designed to communicate with users who are 
not able to provide any more hint than "\emph{Yes, this example works}" or 
"\emph{No, this example doesn't work}". 
We can imagine cases where the user is a bit more skilled than that and can 
provide answers such as  
(a) "\emph{This example $e$ doesn't work because there is something wrong on 
	the variables in this set $Y$}" or 
(b) "\emph{This example $e$ doesn't work because it violates this constraint $c$}" or 
(c) "\emph{This example $e$ doesn't work: here is the set of all the constraints 
	that it violates}". 
\qat can easily  be adapted to these more informative types of answers. 
In the case of (a) we just have to skip the call to \fscope. 
In the case of (b), we can both skip  \fscope and \fc.  
The case (c)  corresponds to the matchmaker agent described in \cite{frewalCP98}. 
The more informative the query, the more dramatic the decrease in 
number of queries needed to find the right constraint network.

\subsection{Illustration  example}

\newcommand{\sumeq}[3]{{\tt \sum_{#1#2}^{=#3}}}
\newcommand{\sumneq}[3]{{\tt \sum_{#1#2}^{\neq#3}}}

We illustrate the behavior of \qat and its two sub-procedures \fscope and \fc 
on  a simple example. 
Consider the  variables $X_1, \ldots, X_5$ with domains $\{-10..10\}$, 
the language $\Gamma=\{\leq,\neq,\sumneq{}{}{}\}$, and 
the basis $B=\{\leq_{ij}, \geq_{ij},  \neq_{ij}\mid i,j\in 1..5,i<j\}\cup
\{\sumneq{i}{j}{k}\mid i,j,k\in 1..5, i< j\neq k\neq i\}$, 
where $\leq_{ij}$ is the constraint $X_i\leq X_j$, 
$\geq_{ij}$ is $X_i\geq X_j$, $\neq_{ij}$ is $X_i\neq X_j$, 
and $\sumneq{i}{j}{k}$ is  $X_i+X_j\neq X_k$.\footnote{Note that $\geq_{ij}$ denotes $\leq_{ji}$.}
The  target network is $T=\{=_{15},<_{23}\sumneq{2}{3}{4}\}$.

Suppose that the first example $e_1$ 
generated in line \ref{qa:gen-sol} of \qat is
$\{X_1=0,X_2=1,X_3=2,X_4=3,X_5=4\}$, denoted by $(0,1,2,3,4)$. 
The query is proposed to the user in  line \ref{qa:ask} of \qat and the user replies \no 
because the constraints $=_{15}$ and  $\sumneq{2}{3}{4}$ are violated. 
As a result, $\fscope(e_1,\emptyset,\{X_1\ldots X_5\})$ is called in  line \ref{qa:findc} of \qat.

\subsubsection*{Running \fscope}\label{sec:example-fs}

\begin{table}[tb]
	\caption{\fscope on the example $(0,1,2,3,4)$}\label{table:ex}
	\begin{center}
		\begin{tabular}[h]{|l|l|l|c|c|}
			\hline
			call & $R$ & $Y$ & ASK & return \\
			\hline
			0 & $\emptyset$ & $X_1, X_2, X_3, X_4, X_5$ & $\times$  &$X_2,X_3, X_4$ \\
			1 & $X_1, X_2, X_3$ &  $X_4, X_5$ & \yes & $X_4$ \\
			1.1 & $X_1, X_2, X_3, X_4$ &  $X_5$&  \no &  $\emptyset $\\
			1.2 & $X_1, X_2, X_3 $ &  $X_4$ & $\times$  & $X_4$\\
			2 & $X_4$ & $X_1, X_2, X_3$ & $\times$ & $ X_2,X_3$\\
			2.1 & $X_1, X_2, X_4$ & $X_3$ & \yes & $X_3$ \\
			2.2 & $X_3, X_4$ & $X_1,X_2$ & \yes & $X_2$ \\
			2.2.1 & $X_1, X_3, X_4$ & $X_2$ & $\times$ & $X_2$ \\
			\hline
		\end{tabular}
	\end{center}
\end{table}

The trace of the execution of $\fscope(e_1,\emptyset,\{X_1\ldots X_5\})$ 
is displayed in Table~\ref{table:ex}. 
Each row corresponds to a call to \fscope. 
Queries are always on the variables in $R$. 
'$\times$' in the column $ASK$ means that the question is skipped 
because $\kk(e_1[R])=\emptyset$. 
This happens when $R$ is of size less than 2 (the smallest constraints in $B$ are binary) 
or because a (positive) query has already been asked 
on $e_1[R]$ and $\kk(e_1[R])$ has been emptied. 

\begin{itemize}
	\item The initial call (call-0 in  Table \ref{table:ex}) does not ask the query 
	because $R=\emptyset$ and $\kk(e_1[\emptyset])=\emptyset$. 
	$Y$ is split in two sets $Y_1=\{X_1, X_2, X_3\}$ and $Y_2=\{X_4, X_5\}$. 
	Line \ref{qa:elu:kappa1} detects that 
	$\kk(e_1[X_1,X_2,X_3])$ and $\kk(e_1[X_1,X_2,X_3,X_4,X_5])$ are 
	different (e.g., $\geq_{25}$ is still in $B$), 
	so the recursive call call-1 is performed. 
	
	\item Call-1: $R=\{X_1, X_2, X_3\}$ (i.e., the $R\cup Y_1$ of call-0) 
	and $Y=\{X_4, X_5\}$ (i.e., the $Y_2$ of call-0). 
	$e_1[X_1, X_2, X_3]$ is classified positive. 
	Hence,  line~\ref{qa:elu:ask} of  \fscope removes all 
	constraints in $\kk(e_1[X_1, X_2, X_3])$ (i.e., $\geq_{12},\geq_{13},\geq_{23}$) from $B$. 
	$Y$ is split in two sets $Y_1=\{X_4\}$ and $Y_2=\{X_5\}$. 
	Again, $\kk(e_1[X_1, X_2, X_3,X_4])$ and $\kk(e_1[X_1,X_2,X_3,X_4,X_5])$ are 
	different in line~\ref{qa:elu:kappa1} ($\geq_{25}$ is still in $B$), so call-1.1 is performed. 
	
	\item Call-1.1:  $e_1[R]$ is classified negative. The empty set is returned 
	in line \ref{qa:elu:emptyset} of call-1.1. 
	We are back to call-1. 
	Line \ref{qa:elu:kappa2} of call-1 detects that 
	$\kk(e_1[X_1,X_2,X_3])$ and $\kk(e_1[X_1,X_2,X_3,X_4,X_5])$ are 
	different, so call-1.2 is performed in line \ref{qa:elu:call2} of call-1. 
	
	\item Call-1.2: $R=\{X_1, X_2, X_3\}$ (i.e., the $S_1\cup Y_1$ of call-1) 
	and $Y=\{X_4\}$ (i.e., the $Y_2$ of call-1). 
	Call-1.2 does not ask the query because $\kk(e_1[X_1, X_2, X_3])$ is 
	already empty (see call-1). 
	In  line \ref{qa:elu:singleton}, call-1.2  detects that $Y$ is a singleton and returns $\{X_4\}$. 
	We are back to call-1. 
	In line \ref{qa:elu:end}, call-1 returns $\{X_4\}$ one level above in the recursion. 
	We are back to call-0. 
	As $\kk(e_1[X_4])$ and $\kk(e_1[X_1,X_2,X_3,X_4,X_5])$ are different, 
	we go to call-2. 
	
	\item Call-2: The query $ASK(e_1[X_4])$ is not asked because $\kk(e_1[X_4])$ is empty. 
	$Y$ is split in two sets $Y_1=\{X_1,X_2\}$ and $Y_2=\{X_3\}$. 
	As $\kk(e_1[X_1,X_2,X_4])$ and $\kk(e_1[X_1,X_2,X_3,X_4])$ are different ($\geq_{34}$ is still in $B$), 
	we go to call-2.1. 
	
	\item Call-2.1: $e_1[X_1,X_2,X_4]$ is classified positive. 
	\fscope removes the constraints in $\kk(e_1[X_1,X_2,X_4])$ from $B$ and returns the singleton $\{X_3\}$. 
	We 	are back to call-2. 
	As $\kk(e_1[X_3,X_4])$ and $\kk(e_1[X_1,X_2,X_3,X_4])$ are different, 
	we go to call-2.2. 
	
	\item Call-2.2:  $e_1[X_3,X_4]$ is classified positive. 
	\fscope removes the constraints in $\kk(e_1[X_3,X_4])$ from $B$. 
	$Y$ is split in two sets $Y_1=\{X_1\}$ and $Y_2=\{X_2\}$. 
	As $\kk(e_1[X_1,X_3,X_4])$ and $\kk(e_1[X_1,X_2,X_3,X_4])$ are different ($\sumneq{2}{3}{4}$ is still in $B$), 
	we go to call-2.2.1. 
	
	\item Call 2.2.1 does not ask the query because $\kk(e_1[X_1, X_3, X_4])$ is 
	empty. (Binary constraints have been removed by  former \yes answers 
	and there is no ternary constraint on $\{X_1, X_3, X_4\}$ that is violated by $e_1$.) 
	As  $Y$ is a singleton. 
	Call-2.2.1  returns $\{X_2\}$. We are back to call-2.2. 
	
	\item
	Line \ref{qa:elu:kappa2} of call-2.2 detects that 
	$\kk(e_1[X_2,X_3,X_4])=\kk(e_1[X_1,X_2,X_3,X_4])$ because 
	all  constraints between $X_1$ and $X_2,X_3,X_4$ that were in $\kk(e_1)$  
	have been removed by \yes answers. Call-2.2.2 is skipped 
	and $\emptyset$ is added to $\{X_2\}$ in line \ref{qa:elu:end} of call-2.2. 
	$\{X_2\}$ is returned to call-2. 
	Call-2 returns $\{X_2,X_3\}$, and call-0 returns $\{X_2,X_3,X_4\}$. 
	Line  \ref{qa:findc} of \qat then calls 
	\fc  with $e_1[X_2, X_3, X_4]=(1,2,3)$ and $Y=\{X_2, X_3, X_4\}$. 
\end{itemize}

\subsubsection*{Running \fc}\label{sec:example-fc}

The trace of the execution of 
$\fc((1,2,3), \{X_2, X_3, X_4\})$ is  displayed in Table~\ref{table:ex:fc}. 
Each row reports  the results of the actions performed after generating 
a new example in line \ref{qa:det:select-ex} of \fc. 
For each of these examples, we report the example generated, its classification, and
the new state of $\Delta$,  $L$, and  $B[X_2, X_3, X_4]$. 
We also specify in which lines of \fc these changes occur. 

\begin{itemize}
	\item 
	Row-0: The example $(1,2,3)$ was not generated in \fc but inherited from \fscope. 
	By definition of \fscope, we know that it is a negative example (denoted by $"(\no)"$ in the table).  
	In  line \ref{qa:det:init} of \fc, $\Delta$ is initialized to the set of constraints 
	from $B$ having scope $\{X_2, X_3, X_4\}$, that is 
	$\{\sumneq{2}{3}{4}, \sumneq{2}{4}{3}, \sumneq{3}{4}{2}\}$, and then in 
	line \ref{qa:det:join} these constraints are   joined with $\sumneq{2}{3}{4}$, 
	the only constraint in  $\kd((1,2,3))$. 
	At this point the learned network $L$ is still empty because \fscope did not modify it. 
	$B[X_2, X_3, X_4]$ contains all the  constraints from the original $B$ 
	with scope included in $\{X_2, X_3, X_4\}$, 
	except $\geq_{23}, \geq_{24}$, and $\geq_{34}$, which were discarded during 
	call-1, call-2.1 and call-2.2 of \fscope, respectively. 
	
	\begin{table}[tb]
		\caption{\fc}\label{table:ex:fc}
		\begin{center}
			\begin{tabular}{|l|l|l|l|l|l|}
				\hline
				& example & ASK &  $\Delta$ &  $L$  & $B[X_2,X_3,X_4]$\\
				& (line \ref{qa:det:select-ex}) 
				& (line \ref{qa:det:ask}) 
				&  (lines \ref{qa:det:init}-\ref{qa:det:join}, \ref{qa:det:cleanBD}, and \ref{qa:det:join2})  
				&  (line \ref{qa:det:return}) 
				& (lines \ref{qa:det:return} and \ref{qa:det:cleanBD})\\
				\hline
				
				\multicolumn{6}{|c|}{$\fc((1,2,3),\{X_2,X_3,X_4\})$}  \\
				\hline
				
				\multirow{2}{*}{0.}
				& \multirow{2}{*}{$(1,2,3)$} &  \multirow{2}{*}{(\no)}  &   $\sumneq{2}{3}{4}$
				&   \multirow{2}{*}{$\emptyset$} 
				& $\sumneq{2}{3}{4},\sumneq{2}{4}{3},\sumneq{3}{4}{2}$ \\
				
				& &  
				&   $\sumneq{2}{3}{4}\land\sumneq{2}{4}{3},\sumneq{2}{3}{4}\land \sumneq{3}{4}{2}$   & 
				& $\leq_{23},\leq_{24},\leq_{34},  \neq_{23},\neq_{24},\neq_{34}$  \\
				\hline
				
				\multirow{2}{*}{1.}
				&  \multirow{2}{*}{$(2,3,1)$} &  \multirow{2}{*}{\yes}
				&   \multirow{2}{*}{$\sumneq{2}{3}{4},\sumneq{2}{3}{4}\land \sumneq{3}{4}{2}$} 
				&  \multirow{2}{*}{$\emptyset$} 
				& $\sumneq{2}{3}{4},\sumneq{3}{4}{2}$ \\
				
				& &  &  & & $\leq_{23}, \neq_{23}, \neq_{24}, \neq_{34}$\\
				\hline
				
				2.  
				&  $(3,2,1)$ & \no &  unchanged   &   $\emptyset$   & unchanged \\
				\hline
				\hline
				
				\multicolumn{6}{|c|}{$\fc((3,2),\{X_2,X_3\})$}  \\
				\hline
				
				3.
				&    $(3,2)$ & (\no) &  $ \leq_{23},<_{23}$ & $\emptyset$ & unchanged \\
				\hline
				
				4.
				& $(1,1)$ & \no & $<_{23}$ & $<_{23} $
				& $\sumneq{2}{3}{4},\sumneq{3}{4}{1}, \neq_{24}, \neq_{34}$ \\
				\hline
				\hline
				
				\multicolumn{6}{|c|}{back to $\fc((1,2,3),\{X_2,X_3,X_4\})$}  \\
				\hline
				
				5.
				& $(1,2,-1)$ & \yes &  $\sumneq{2}{3}{4} $  & $<_{23},\sumneq{2}{3}{4}$
				& $\neq_{24}, \neq_{34}$  \\
				\hline
			\end{tabular}
		\end{center}
	\end{table}

	\item 
	Row-1: In  line \ref{qa:det:select-ex}, \fc  generates the example $(2,3,1)$, 
	satisfying some constraints from $\Delta$ but not all. 
	$(2,3,1)$ is classified positive in line \ref{qa:det:ask}. 
	Hence, the violated conjunction  $\sumneq{2}{3}{4}\wedge \sumneq{2}{4}{3}$ 
	is removed from $\Delta$ and all violated constraints in $B$  
	(i.e., $\sumneq{2}{4}{3}$,  $\leq_{24}$, and $\leq_{34}$) are 
	removed (line \ref{qa:det:cleanBD}). 
	$L$ remains unchanged.

	\item
	Row-2: \fc generates the example $(3,2,1)$, which is classified negative. 
	The call to  \fscope in line \ref{qa:det:findsc} returns $S=\{X_2,X_3\}$. 
	Line \ref{qa:det:findc} then recursively calls  \fc on the scope $\{X_2,X_3\}$.   
	
	\item
	Row-3: 	The example $(3,2)$ is known to be negative without asking. 
	Lines \ref{qa:det:init}-\ref{qa:det:join} initialize $\Delta$ to the set of constraints 
	in $B_{\{X_2,X_3\}}$ and then join them to those rejecting the example. 
	(Note that $<_{23}$ is a shortcut for $\leq_{23}\land\neq_{23}$.)   
	$L$ and $B$ remain unchanged. 
	
	\item 
	Row-4: \fc generates the example $(1,1)$, which is classified negative. 
	The call to  \fscope in line \ref{qa:det:findsc} returns the same scope $S=\{X_2,X_3\}$ 
	because $B$ does not contain any smaller arity constraints. 
	Line \ref{qa:det:join2} reduces $\Delta$ to a the singleton $<_{23}$. 
	As a result, the next loop of \fc cannot generate any new example in line \ref{qa:det:select-ex}. 
	Line \ref{qa:det:return} adds $<_{23}$ to $L$ and removes all the constraints with scope 
	$\{X_2,X_3\}$ from $B$. This subcall to \fc exits. 
	
	\item
	Row-5:  We are back to the original call to \fc with the same $\Delta$  as in row-2.  
	Line \ref{qa:det:select-ex} must  generate an  example accepted by $L$ and violating 
	part of $\Delta$. It generates  $(1,2,-1)$, which is positive. 
	The violated conjunction  $\sumneq{2}{3}{4}\wedge \sumneq{3}{4}{2}$ 
	is removed from $\Delta$ and 
	$\sumneq{3}{4}{2}$ is removed from $B$  (line \ref{qa:det:cleanBD}). 
	The next loop of \fc cannot generate any new example in line \ref{qa:det:select-ex} 
	because $\Delta$ is now a singleton. 
	Line \ref{qa:det:return} adds $\sumneq{2}{3}{4}$ to $L$ and removes all 
	the constraints with scope $\{X_2,X_3,X_4\}$ from $B$. 
	\fc exits. 
\end{itemize}

%
%

\subsection{Theoretical analysis}\label{sec:qa:theo}

We first show that \qat is a correct algorithm to learn a constraint 
network equivalent to a target network 
that can be specified within a  given basis. 
We  prove that \qat is sound, complete, and terminates. 

\begin{proposition}[Soundness]\label{prop:sound}
	Given a basis $B$  and 
	a target network $T\subseteq B$, 
	the network $L$ returned by \qat is such that 
	$sol(T)\subseteq sol(L)$. 
\end{proposition}

\proof
Suppose there exists $e_1\in sol(T)\setminus sol(L)$. 
Hence, there exists at least one scope on which \qat has learned 
a conjunction of constraints rejecting $e_1$. 
Let us consider the first such conjunction $c^*$ learned by \qat, and let us denote  
its scope by $Y$. 
By assumption, $c^*$ contains an elementary constraint $c_1$ rejecting $e_1$. 
The only place where we add a conjunction of constraints to $L$ is 
line \ref{qa:det:return} of \fc. 
This conjunction has been built by join operations in 
lines \ref{qa:det:join} and \ref{qa:det:join2} of \fc. 
By construction of \fscope, 
$e_1[Y]$ is rejected by a constraint with scope $Y$ in $T$ 
and by none of the constraints  on subscopes of $Y$ in $T$ 
when the join operation  in line \ref{qa:det:join} of \fc is executed. 
By construction of \fc, 
the join operations in line \ref{qa:det:join2} of \fc are executed 
for and only for  $e'_Y$ generated in this call to \fc  
that are rejected by a constraint with scope $Y$ in $T$ 
and by none of the constraints on subscopes of $Y$. 
As a result, $\Delta$ contains all minimal 
conjunctions of elementary constraints from $B_Y$ 
that reject $e_1[Y]$ and all $e'_Y$ generated in this call to \fc 
that are rejected by a constraint of scope $Y$ in $T$ 
and by none of the constraints on subscopes of $Y$. 
One of those minimal conjunctions is necessarily a subset of the conjunction in $T$. 
In line \ref{qa:det:return}, when  we put one of these conjunctions in $L$, 
they are all equivalent wrt $L$ because line \ref{qa:det:select-ex} 
could not produce an example $e'_Y$ violating some conjunctions 
from $\Delta$ and satisfying the others. 
As  scope $Y$ is, by assumption, the first scope on which \qat learns 
a wrong conjunction of constraints, we deduce that all conjunctions 
in $\Delta$ are  equivalent wrt to $T$. As a consequence,  none can contain $c_1$. 
Therefore,  adding one of them  to $L$ 
cannot reject $e_1$. 
\qed

\begin{proposition}[Completeness]\label{prop:complete}
	Given a basis $B$  and 
	a target network $T\subseteq B$, 
	the network $L$ returned by \qat is such that 
	$sol(L)\subseteq sol(T)$. 
\end{proposition}

\proof
Suppose there exists $e_1\in sol(L)\setminus sol(T)$ when \qat terminates. 
Hence, there exists an elementary  constraint $c_1$ in $B$ that rejects $e_1$,  
and $c_1$ belongs to $c^*$, the conjunction of the constraints in  $T$ with same 
scope as $c_1$. 
%
%
%
%
The only way for \qat to terminate is  
line \ref{qa:converge} of \qat. 
This means that in line~\ref{qa:gen-sol}, \genex was not able 
to generate an example $e_Y$ accepted by $L[Y]$ 
and rejected by $B[Y]$. 
Thus, $c_1$ is not in $B$ when \qat terminates, 
otherwise the projection $e_1[Y]$ 
of $e_1$ on any $Y$ containing $var(c_1)$ would have been such an example. 
We know that $c_1\in T$, so $c_1$ was in $B$ before starting \qat. 
Constraints can be removed from $B$ in 
line  \ref{qa:remove}  of \qat, 
line \ref{qa:elu:ask}   of  \fscope, and
lines \ref{qa:det:return} and \ref{qa:det:cleanBD} of  \fc. 
In line  \ref{qa:remove}  of \qat, line \ref{qa:elu:ask}   of \fscope, 
and line \ref{qa:det:cleanBD} of \fc, 
a constraint $c_2$ is removed from $B$ because it rejects a positive example. 
This removed constraint $c_2$ cannot be  $c_1$ because  $c_1$  
belongs to $T$, so it cannot reject a 
positive example. 
In line \ref{qa:det:return} of  \fc, 
all (elementary)  constraints with scope $Y$ are removed from $B$. 
Let us see if one of them could be our $c_1$. 
Given an elementary constraint $c_2$ with scope $Y$ that is removed from 
$B$ in line \ref{qa:det:return} of  \fc,  either $c_2$ is still 
appearing in one conjunction of $\Delta$ when \fc terminates, or not. 
Thanks to lines \ref{qa:det:nil} and \ref{qa:det:pick}, we know that 
$L\cup\{c_Y\}\models \Delta$. Thus, if $c_2$ is  in one of the conjunctions of $\Delta$, 
then $L\models c_2$ after the execution of 
line \ref{qa:det:return}, the only line where \fc can terminate. 
Thus, $c_2$ cannot be $c_1$ because by assumption $c_1$ rejects $e_1$, which 
itself is accepted by $L$. 
If  $c_2$ is  not in any of the conjunctions of 
$\Delta$ when \fc terminates, these conjunctions  
must have been removed in line \ref{qa:det:cleanBD} 
or in line \ref{qa:det:join2}, the two places where $\Delta$ is modified. 
Let us denote by $\widehat{c_2}$ a conjunction in $\Delta$ composed of $c_2$ and a subset of $c^*$. 
It necessarily exists at the first execution of the loop in line \ref{qa:det:loop} 
because $c_2\in B$ and line \ref{qa:det:join} either keeps $c_2$ (if $c_2$ is violated by $e$) , 
or joins it with elements of $c^*$ (if  $c_2$ is satisfied by $e$). 
Line \ref{qa:det:join2} is executed after a negative query $e'_Y$. 
If $c_2$ rejects $e'_Y$, all the conjunctions containing it remain in $\Delta$. 
If $c_2$ is satisfied by $e'_Y$, there necessarily exists a conjunction  
in $\kd(e'_Y)$ which is a subset of the conjunction $c^*$ because 
\qat is sound (Proposition \ref{prop:sound}). 
$\widehat{c_2}$ is joined with this subset. 
Thus, $\Delta$ still contains a conjunction composed of $c_2$ and a subset of $c^*$. 
Each time a negative example will be generated, this subset will 
either stay in $\Delta$ or be joined with another subset of $c^*$. 
As a result,  line \ref{qa:det:join2} cannot remove all conjunctions  
composed of  $c_2$ and a subset of $c^*$. 
These conjunctions $\widehat{c_2}$ must then  have been removed in line \ref{qa:det:cleanBD} 
because they were rejecting the  example $e'_Y$ classified positive in 
line~\ref{qa:det:ask}. 
These conjunctions  can be removed only if $c_2$  rejects $e'_Y$ because the rest 
of the conjunction is a subset of $c^*$. 
Again $c_2$ cannot be $c_1$ because $c_1$ 
cannot reject positive examples.  
Therefore, $c_1$ cannot reject an example accepted by $L$, 
which proves that $sol(L)\subseteq sol(T)$. 
\qed

\begin{proposition}[Termination]\label{prop:termination}
	Given a basis $B$  and 
	a target network $T\subseteq B$, 
	\qat terminates. 
\end{proposition}

\proof
Each execution of the loop in line \ref{qa:loop} of \qat either 
executes line~\ref{qa:remove} of  \qat 
or enters  \fc. 
By construction of $e_Y$ in line \ref{qa:gen-sol} of \qat 
we know that $\kk(e_Y)$ is not empty. 
Hence, in line~\ref{qa:remove} of  \qat, 
$B$ strictly decreases in size. 
By definition of  \fscope, the set $Z$ returned by \fscope 
is such that there exists a constraint $c$ with $var(c)=Z$ in $B$ rejecting $e_Y$. 
Thus,  $\kk(e_Y[Z])$ is not empty. 
As a result, each time \fc is called, $B$ strictly decreases in size 
because  \fc always executes line~\ref{qa:det:return} before exiting. 
Therefore, at each execution of the loop  in line~\ref{qa:loop} of 
\qat, $B$ strictly decreases in size.  As $B$ has 
finite size, we have termination. 
\qed

\begin{theorem}[Correctness]\label{prop:correct}
	Given a basis $B$  and 
	a target network $T\subseteq B$, \qat returns a 
	network $L$ such that $sol(L)= sol(T)$. 
\end{theorem}

\proof
Correctness immediately follows from Propositions \ref{prop:sound}, \ref{prop:complete}, and \ref{prop:termination}. 
\qed\\

We analyze the complexity of \qat in terms of the number of queries it
can ask of the user. Queries are proposed to the user in line
\ref{qa:ask} of \qat, line  \ref{qa:elu:ask} of   \fscope and
line \ref{qa:det:ask} of \fc.

\begin{proposition}\label{prop:elu}
	Given a vocabulary $(X,D)$,  a basis $B$, 
	a target network $T$,  and 
	an example $e_Y\in D^Y$ rejected by $T$, 
	\fscope uses $O(|S|\cdot\log|Y|)$ queries to
	return the scope $S$ of one of the constraints of 
	$T$ violated by $e_Y$. 
\end{proposition}

\proof
Let us first consider a version of \fscope that would execute  lines 
\ref{qa:elu:call1} and \ref{qa:elu:call2} unconditionally. 
That is, a version without 
the tests in lines \ref{qa:elu:kappa1} and \ref{qa:elu:kappa2}. 
\fscope  is a recursive algorithm that asks at most one query
per call (line \ref{qa:elu:ask}). Hence, the number of queries is
bounded above by the number of nodes of the tree of recursive calls to 
\fscope. 
We  show that a leaf node  is either on a branch that leads to the
elucidation of a variable in the scope $S$ that will be returned, or is
a child of a node of such a branch. 
By construction of \fscope, we observe that \no answers  
to the query in line \ref{qa:elu:ask} always occur in leaf calls and that the only
way for a leaf call to return the empty set is to have received
a \no answer  to its query (line \ref{qa:elu:emptyset}). 
Let $R_{child},Y_{child}$ be the values of the parameters $R$ and $Y$ for a
leaf call with a \no answer, 
and $R_{parent},Y_{parent}$ be the values of the parameters $R$ and $Y$
for its parent call in the recursive tree. 
We know that  $S\nsubseteq R_{parent}$ because the parent call 
necessarily received a \yes answer. 
Furthermore, from 
the \no answer to the query $ASK(e[{R_{child}}])$, we know that
$S\subseteq R_{child}$. 
Consider first the case where the leaf is the left child of the parent
node. 
By construction, $R_{child}\subsetneq R_{parent}\cup Y_{parent}$.  
As a result, $Y_{parent}$ intersects $S$, and the parent node is on a
branch that leads to the elucidation of a variable in $S$. 
Consider now the case where the leaf is the right child of the parent
node. 
As we are on a leaf, if the test of line \ref{qa:elu:doweask} is false 
(i.e., $\kk(e[R_{child}])=\emptyset$), we 
necessarily exit from \fscope through line \ref{qa:elu:singleton}, 
which means that this node is the end of a branch leading 
to a variable in $S$. 
If the test of line \ref{qa:elu:doweask} is true 
(i.e., $\kk(e[R_{child}])\neq\emptyset$), 
we are guaranteed that the left child of the parent node returned 
a non-empty set, otherwise $R_{child}$ would be equal to $R_{parent}$ 
and we know that  $\kk(e[R_{parent}])$ has been emptied 
in line \ref{qa:elu:cleanB} as it received a \yes answer. 
Thus,  the parent node is on a branch to a leaf that
elucidates a variable in $S$. 

We have proved that every leaf is either on a branch that elucidates a
variable in $S$ or is a child of a node on such a branch. 
Hence the number of nodes in the tree is at most twice the number of
nodes in branches that lead to the elucidation of a variable from
$S$. 
Branches can be at most $\log|Y|$ long. Therefore the total
number of queries \fscope can ask is at most $2\cdot |S|\cdot \log|Y|$,
which is in $O(|S|\cdot\log|Y|)$. 

Let us come back to the complete version of \fscope, where lines 
\ref{qa:elu:kappa1} and \ref{qa:elu:kappa2} are active. 
The purpose of lines \ref{qa:elu:kappa1} and \ref{qa:elu:kappa2} is only 
to avoid useless calls to \fscope that would return $\emptyset$ anyway. 
These lines do not affect anything else in the algorithm. 
Hence, by adding lines \ref{qa:elu:kappa1} and \ref{qa:elu:kappa2}, 
we can only decrease the number of recursive calls to \fscope. 
As a rsult, we cannot increase the number of queries.  
\qed

%

\begin{theorem}[Complexity]\label{theo:complexity}
	Let $\Gamma$ be a language of bounded-arity relations. 
	\qat learns constraint networks over $\Gamma$ in $O(m\log n+b)$ queries,   
	where $n$ and $m$ are respectively the number of variables  and the number 
	of constraints of the target network, and $b$ is the size of the basis. 
\end{theorem}

\proof
Each time line \ref{qa:ask} of \qat classifies an example as negative,
the scope $var(c)$ of a constraint $c$ from the target network is found in $O(|var(c)|\cdot\log n)$
queries (Proposition~\ref{prop:elu}).  
As the basis only contains
constraints of bounded arity,  $var(c)$ is found in $O(\log n)$ queries. 
Finding $c$  with \fc requires a number of queries in $O(1)$ 
because the size of $\Gamma$ does not depend on the size of the target network. 
Hence, the number of queries
necessary for finding the target network is in $O(m\log n)$. 
Convergence is obtained once the basis is wiped out  of all its constraints 
or those remaining are implied by the learned network $L$.  
Each time  an  example is 
classified positive in line \ref{qa:ask} of \qat or 
line \ref{qa:elu:ask} of \fscope,  
this leads to at least one constraint removal from the basis
because, by construction of \qat and \fscope,  
this example violates at least one constraint from the basis. 
Concerning queries asked in  \fc, their number is in $O(1)$ 
at each call to \fc, and there are no more calls to \fc than 
constraints in the target network because \fc always adds at least one constraint 
to $L$ during its execution (line \ref{qa:det:return}). 
This gives a total number of queries required for convergence 
that is bounded above by the size $b$ of the basis. 
\qed\\

The complexities stated in Theorem \ref{theo:complexity} 
are based on the size of the target network and size of 
the basis. 
The size of the language $\Gamma$ is not considered because 
it has a fixed size, independent on the number of variables in the target network. 
Nevertheless, line \ref{qa:det:join2} of \fc can lead to an increase 
in the size of $\Delta$  up to $2^{|\Gamma|}$. 
By reformulating line \ref{qa:det:select-ex} 
of \fc 
as shown below, we can bound the increase in size of $\Delta$. 
In the following, we use the notation $\Delta_p$ as defined at the very end of 
Section \ref{sec:background}.  

\LinesNotNumbered
\begin{algorithm}[htb]
	
	{...}
	
	\nlset{5bis}	
	choose $e'_Y$ in $sol(L[Y])$ and $\emptyset \subsetneq \kd(e'_Y) \subsetneq\Delta$, 
	minimizing $p$ such that
	$\emptyset \subsetneq\kdp(e'_Y)\subsetneq\Delta_p$ if possible,  
	$\kdp(e'_Y) \subsetneq\Delta_p$ otherwise\label{qa:det:select-ex-v2}\; 
	{...}
	
	\end{algorithm}
	\LinesNumbered
	
	\begin{proposition}
		Given a basis $B$, a target network $T$, and a 
		scope $Y$, the number of queries 
		required by \fc to learn a subset of $B_Y$ equivalent to 
		the conjunction of  constraints of $T$ with scope $Y$ in $T[Y]$  
		is in  $O(|B_Y|+2^{max(|c^*|,|I_{c^*}|)})$, where 
		$c^*$ is the  smallest such conjunction and 
		$I_{c^*}=\{c_i\in B_Y\mid c^*\to c_i\}$.  	
		%
		
	\end{proposition}
	
	\proof
	We first compute the number of queries required to generate 
	$c^{*}$ in $\Delta$, and then the number of queries required 
	to remove all conjunctions of constraints 
	not equivalent to $c^*$ from $\Delta$. 
	
	Let us first prove that line \ref{qa:det:select-ex-v2} of 
	\fc will not stop generating  examples before  $c^*$ is 
	one of the conjunctions in $\Delta$. 
	Let us take as induction hypothesis that  
	when entering a new execution of the 
	loop in line \ref{qa:det:loop}, if $c^*$ is not in $\Delta$, 
	then the set of the conjunctions 
	in $\Delta$ that are included in $c^*$ covers
	the whole set of elementary  constraints from $c^*$. 
	That is, $\bigcup\{sub\in \Delta\mid sub\subset c^*\}=c^*$. 
	The only way to modify $\Delta$ is to ask a query $e'_Y$. 
	If  $e'_Y$ is positive, this means that $c^*$ is satisfied 
	and all its subsets remain in $\Delta$. 
	If $e'_Y$ is negative,  either this is due to a constraint of 
	$T$ on a subscope of $Y$ or not. 
	If it is due to a constraint on a subscope, 
	line \ref{qa:det:findc} is executed and not line \ref{qa:det:join2}, so 
	$\Delta$ remains unchanged. 
	If it is not due to a constraint on a subscope, 
	this  guarantees that at least  
	one elementary constraint of  $c^*$ is  violated, and according 
	to our induction hypothesis, at least one subset of $c^*$, 
	call it $sub_1$, is in $\kd(e'_Y)$. 
	Hence, line \ref{qa:det:join2}  generates a conjunction 
	of $sub_1$ with each of the other subsets of  $c^*$ that are in $\Delta$. 
	As a result, every elementary constraint in $c^*$  belongs 
	to at least one of these conjunctions with $sub_1$ that are uniquely composed
	of elementary constraints from $c^*$. 
	Furthermore, before line \ref{qa:det:join}, by construction, 
	all elementary constraints composing  $c^*$ are in $\Delta$ 
	and line   \ref{qa:det:join} is similar to line  \ref{qa:det:join2}. 
	As a consequence, our induction hypothesis is true. 
	We prove now that as long as $c^*$ is not in $\Delta$, line \ref{qa:det:select-ex-v2} 
	is  able to  generate a query $e'_Y$. 
	By definition, we know that $c^*$ is the smallest conjunction 
	equivalent to the constraint of $T$ with scope $Y$. 
	Thus, no subset of $c^*$ can be implied by any other subset of $c^*$. 
	This  guarantees that there exists an example $e'_Y$ such that 
	one subset $sub_1$ of  $c^*$ is in $\kd(e'_Y)$ and another subset, $sub_2$,  
	is  in $\Delta\setminus \kd(e'_Y)$. 
	$e'_Y$ is a valid query to be generated in line \ref{qa:det:select-ex-v2} 
	and to be asked in line \ref{qa:det:ask}. 
	As a consequence, we cannot exit \fc as long 
	as $c^*$ is not in $\Delta$. 
	
	We now prove that  $c^*$ is in $\Delta$ after  a 
	number of queries linear in $|B_Y|$. 
	We first count the number of  positive queries. 
	Thanks to the condition in line \ref{qa:det:select-ex-v2} of \fc, we 
	know that at least one  elementary constraint $c_i$ of $B_Y$ is violated by 
	the query. Thus, all the conjunctions containing $c_i$ are removed 
	from $\Delta$ in line \ref{qa:det:cleanBD}, and no conjunction containing $c_i$
	will  be able to come again in $\Delta$. 
	As a result, the number of positive queries is bounded above by $|B_Y|$. 
	Let us now count the number of negative queries. 
	A query can be negative because of a constraint on a subscope of $Y$ or because 
	of $c^*$. 
	If because of a subscope we do not count it in the cost of learning $c^*$. 
	If because of $c^*$, we saw that there exists a subset $sub_1$ of $c^*$ in $\kd(e'_Y)$. 
	Line \ref{qa:det:join2}  generates a conjunction 
	of $sub_1$ with each of the other subsets of  $c^*$ that are in $\Delta$. 
	Before the joining operation, either $sub_1$ is included in the largest 
	subset $maxsub$ or not. 
	If $sub_1$ is included in $maxsub$, 
	then $maxsub$ also belongs to $\kd(e'_Y)$ and it  
	produces a larger subset by joining with any other non-included subset of $c^*$. 
	If $sub_1$ is not included in $maxsub$, they are necessarily  joined 
	together, generating again a subset strictly larger than $maxsub$. 
	Thus, the number of queries that are negative because of $c^*$ is 
	bounded above by  $|c^*|$. 
	Therefore, the number of queries necessary to have  $c^*$  in $\Delta$ 
	is in $O(|B_Y|)$. 
	
	Once  $c^*$  has been generated, it will remain in $\Delta$  until the 
	end of this call to \fc because it can be removed neither by a positive 
	query (it would not be in $\kd(e'_Y)$) nor by a negative (either it is 
	in the $\kd(e'_Y)$ or a subconstraint is found and $\Delta$ is 
	not modified). 
	
	We now show that  the  number of queries required 
	to remove all conjunctions of constraints 
	not equivalent to $c^*$ from $\Delta$ is in $O(|B_Y|+2^{max(|c^*|,|I_{c^*}|)})$. 
	We first have to prove that 
	once a conjunction $rem$ has been removed from $\Delta$, 
	it will never come back in $\Delta$ by some join operation. 
	The conjunction $rem$ can come back in $\Delta$ if and only if 
	there exist $a$ and $b$ in $\Delta$ such that $rem=a\land b$. 
	If $rem$ was removed due to a positive query $e'_Y$, 
	then $rem$ was in $\kd(e'_Y)$ and then, either $a$ or $b$ 
	was in $\kd(e'_Y)$ too. Thus, $a$ or $b$ has been removed from $\Delta$ at the 
	same time as $rem$, which contradicts the assumption that $rem$ came back due to
	the join of $a$ and $b$. 
	%
	%
	If $rem$ was removed due to a negative query $e'_Y$, 
	then $rem$ was not in $\kd(e'_Y)$ and then, none of $a$ and $b$ 
	were in $\kd(e'_Y)$. 
	$a$ and $b$ have thus both been joined with other elements of $\kd(e'_Y)$ and 
	have  disappeared from $\Delta$ at the 
	same time as $rem$. This again contradicts the assumption.

	We are now ready to show that  all conjunctions not equivalent to $c^*$ are 
	removed from $\Delta$ in $O(|B_Y|+2^{max(|c^*|,|I_{c^*}|)})$ queries. 
	For that, we first prove that all conjunctions not implied by 
	$c^*$ are removed from $\Delta$ in $O(|B_Y|+2^{|c^*|})$  queries. 
	As long as there exists a conjunction $nimp$  in $\Delta$ such that 
	$c^*\not\models nimp$,  line \ref{qa:det:select-ex-v2} can generate a query $e'_Y$ 
	with $p\leq|c^*|$. If 
	$\emptyset\subsetneq \kdp(e'_Y)\subsetneq \Delta_p$ cannot be satisfied 
	for any $p\leq|c^*|$, then
	there necessarily exists an $e'_Y$ (satisfying $c^*$ and violating $nimp$) 
	with $\kdc(e'_Y)=\emptyset\subsetneq \Delta_{|c^*|}$  
	and $nimp\in\kd(e'_Y)$, 
	otherwise we would have $c^*\models nimp$.
	%
	%
	%
	As a result,  line \ref{qa:det:select-ex-v2} can 
	never return a query $e'_Y$ with $p>|c^*|$ 
	if there exists $nimp$ in $\Delta$ such that $c^*\not\models nimp$. 
	Suppose first that $ASK(e'_Y)=yes$. 
	By construction of $e'_Y$, we know that at least one  elementary constraint $c_i$ 
	of the initial $B_Y$ (line \ref{qa:det:init}) is violated by $e'_Y$. 
	Thus, all the conjunctions containing $c_i$ are removed from $\Delta$ 
	and the number of positive queries is bounded above by $|B_Y|$. 
	Suppose now that $ASK(e'_Y)=no$. 
	By construction of $e'_Y$, 
	we know that $\Delta_p\setminus\kdp(e'_Y)$ is not empty for some $p\leq|c^*|$, 
	and all these conjunctions in $\Delta_p\setminus\kdp(e'_Y)$ 
	disappear from $\Delta_p$ in line \ref{qa:det:join2} 
	because they are joined with other conjunctions of $\kd(e'_Y)$. 
	Hence, the number of negative queries is bounded above by the number 
	of possible conjunctions in $\Delta_{|c^*|}$, which is in $O(2^{|c^*|})$.

	Once all the conjunctions not implied by $c^*$ have been removed from $\Delta$,  
	$\Delta$ only contains $c^*$ and conjunctions 
	included in the set $I_{c^*}$ of elementary constraints implied by $c^*$. 
	We show that removing from $\Delta$ all conjunctions implied by $c^{*}$
	is performed in $O(2^{|I_{c^*}|})$ queries. 
	As all conjunctions remaining in $\Delta$ are implied by $c^*$, all queries 
	will be negative.  
	By construction of such a negative query $e'_Y$,  
	we know that $\Delta\setminus\kd(e'_Y)$ is not empty. 
	All these conjunctions in $\Delta\setminus\kd(e'_Y)$ 
	disappear from $\Delta$ in line \ref{qa:det:join2} 
	because they are joined with other conjunctions of $\kd(e'_Y)$. 
	Thus, each query  removes at least one 
	element from $\Delta$, which is a subset of $\{c^*\}\cup 2^{I_{c^*}}$. 
	As a result, the number of such queries is in $O(2^{|I_{c^*}|})$. 
	\qed

	\begin{corollary} 
		\label{coro:linear}
		Given a basis $B$, a target network $T$, and a 
		scope $Y$ such that $B_Y$ contains a constraint $c^*$  equivalent to 
		the conjunction of  constraints of $T$ with scope $Y$ 
		and there does not exist any $c$ in $B_Y$ such that $c^*\to c$,  
		\fc  returns $c^*$ in  $O(|B_Y|)$ queries, which is included in $O(|\Gamma|)$. 
	\end{corollary}
	
	The good news brought by Corollary \ref{coro:linear} are that 
	despite the join operation required in \fc to deal with non-normalized 
	networks, \qat is linear in the size of the language $\Gamma$ when 
	the target network is normalized and  $\Gamma$ does not contain constraints subsuming others.

	\section{Learning Simple Languages}
	\label{sec:formal}

	
	The performance of \qat (in terms of the number of queries submitted to the user) crucially depends on the nature of the relations in the language $\Gamma$. Some constraint languages are intrinsically harder to learn than others, and there may exist languages that are easy to learn using a specialized algorithm but difficult to learn using \qat.
	
	Determining precisely how \qat fares when compared with an optimal learning algorithm (that uses partial queries) on a given language $\Gamma$ is in general a very difficult question. However, if $\Gamma$ is simple enough then a complete analysis of the efficiency of \qat is possible. In this section, we focus on constraint languages built from the elementary relations $\{ = , \neq, >\}$ and systematically compare \qat with optimal learning algorithms. We will measure the number of queries as a function of the number $n$ of variables; our analysis only assumes that the example $e_Y$ generated in line \ref{qa:gen-sol} of \qat\ is \emph{complete} (i.e., $Y = X$) and is a solution of $L$ that \emph{maximizes} the number of violated constraints in the basis $B$.
	
	The next Theorem summarizes our findings. For the sake of readability, its proof is delayed at the end of the section.

	
	\begin{theorem}
		\label{thm:languages}
		Let $\Gamma \subseteq \{=, \neq, >\}$ be a non-empty constraint language over a finite domain $D \subset \mathbb{Z}$, $|D| > 1$. The following holds:
		\begin{itemize}
			\item If $|D| = 2$, then \qat learns networks over $\Gamma$ in $\Theta(n \log n)$ queries in the worst case. This is asymptotically optimal, except for $\Gamma = \{ > \}$ for which the optimum is $\Theta(n)$.
			\item If $|D| > 2$, then in the worst case \qat learns networks over $\Gamma$ in 
			\begin{itemize}
				\item[(i)] $\Theta(n \log n)$ queries if $\Gamma = \{ = \}$, which is asymptotically optimal, and
				\item[(ii)] $\Theta(n^2 \log n)$ queries otherwise, while the optimum is $\Theta(n^2)$.
			\end{itemize}
		\end{itemize}
	\end{theorem}
	
	Note that for all these languages, the asymptotic number of queries made by \qat differs from the best possible by a factor that is at most logarithmic.
	
	The proof of Theorem~\ref{thm:languages} is based on the following six lemmas. The first three (Lemmas~\ref{lem:rectangularity}, \ref{lem:lb-equal} and \ref{lem:lb-neq}) derive unconditional lower bounds on the number of queries necessary to learn certain constraint languages from a simple counting argument. Lemmas~\ref{lem:ub-equal}, \ref{lem:ub-equalsup} and \ref{lem:algoless} (together with Theorem~\ref{theo:complexity}) will then establish matching upper bounds.
	
	
	\begin{lemma}
		\label{lem:rectangularity}
		Let $\Gamma$ be a constraint language over a finite domain $D \subset \mathbb{Z}$, $|D| > 2$, such that $\{ >, \neq \} \cap  \Gamma \neq \emptyset$. Then, learning constraint networks over $\Gamma$ requires $\Omega(n^2)$ partial queries in the worst case.
	\end{lemma}
	
	\proof
	Let $d_1,d_2,d_3$ be three values in $D$ such that $d_1 > d_2 > d_3$ and $(X,D)$ be a vocabulary with an even number $n$ of variables. Let ${\cal C}_n$ denote the set of all possible solution sets of constraint networks over $\Gamma$ with vocabulary $(X,D)$. For any $(i,j) \in [1,\ldots,n/2] \times [n/2+1,\ldots,n]$ we define the assignment $\phi_{ij} : X \to D$ as follows: 
	\begin{eqnarray*}
		\phi_{ij}(X_q) & = \left\{ \begin{array}{ll} 
			d_1 & \mbox{if } q \in [1,\ldots,n/2] \backslash \{i\} \\
			d_2& \mbox{if } q \in \{i,j\} \\
			d_3 & \mbox{if } q \in [n/2+1,\ldots,n] \backslash \{j\} \end{array}\right.
	\end{eqnarray*}
	Now, let $R$ denote a relation in $\{ >\, , \neq \} \cap \Gamma$ and observe that $R$ contains the three tuples $(d_1,d_2)$, $(d_1,d_3)$, $(d_2,d_3)$ but not the tuple $(d_2,d_2)$. Then, for any subset $S \subseteq {\cal S} = \{ \phi_{ij} \mid (i,j) \in [1,\ldots,n/2] \times [n/2+1,\ldots,n] \}$ the constraint network $C^S = \{ R_{(X_i,X_j)} : \phi_{ij} \notin S\}$ over $\Gamma$ has the property that $sol(C^S) \cap {\cal S} = S$. In particular, for any two distinct sets $S_1, S_2 \subseteq {\cal S}$ we have $sol(C^{S_1}) \neq sol(C^{S_2})$ and hence
	\[
	|{\cal C}_n| \geq |\{ C^S \mid S \subseteq {\cal S} \}| = 2^{|{\cal S}|} = 2^{(n/2)^2}.
	\]
	It follows that learning constraint networks over $\Gamma$ requires $\Omega(n^2)$ partial queries since each query only provides a single bit of information on the target network. 
	\qed
	
	\begin{lemma}
		\label{lem:lb-equal}
		Let $\Gamma$ be a constraint language such that $\{=\} \subseteq \Gamma$. Then, learning constraint networks over $\Gamma$ requires $\Omega( n \log n )$ partial queries in the worst case.
	\end{lemma}
	
	\proof
	In a constraint network over $\{=\}$, all variables of a connected component must be equal. In particular, two constraint networks over $\{=\}$ with the same variable set $X$ are equivalent (i.e. have the same solution set) if and only if the partitions of $X$ induced by the connected components are identical. The number of possible partitions of $n$ objects is known as the $n$th \emph{Bell Number} $C(n)$. It is known that $\log C(n) = \Omega(n \log n)$~\cite{de1970asymptotic}, so this entails a lower bound of $\Omega(n \log n)$ queries to learn constraints networks over $\Gamma$.
	\qed
	
	\begin{lemma}
		\label{lem:lb-neq}
		Let $\Gamma$ be a constraint language over a domain $D \subset \mathbb{Z}$, $|D| = 2$, such that $\{\neq\} \subseteq \Gamma$. Then, learning constraint networks over $\Gamma$ requires $\Omega( n \log n )$ partial queries in the worst case.
	\end{lemma}
	
	\proof
	Since $|D| = 2$ and $\{\neq\} \subseteq \Gamma$, we can simulate an equality constraint $X_i = X_j$ over $D$ by introducing one fresh variable $X_{ij}$ and two constraints $X_i \neq X_{ij}$, $X_{ij} \neq X_j$. It follows that for every set $S^{=}$ of non-equivalent constraint networks over $\{=\}$ with domain $D$, $n$ variables and $O(n)$ constraints, we can construct a set $S^{\neq}$ of non-equivalent constraint networks over $\Gamma$ with $n^* = O(n)$ variables and such that $|S^{\neq}| = |S^=|$. As we have seen in the proof of Lemma~\ref{lem:lb-equal}, $|S^=|$ can be chosen such that $\log |S^=| = \Omega(n \log n)$. In that case, we have $\log |S^{\neq}| = \Omega(n^* \log n^*)$ and the desired lower bound follows.
	\qed
	
	\begin{lemma}
		\label{lem:ub-equal}
		For any finite domain $D \subset \mathbb{Z}$ with $|D| \geq 2$, \qat learns constraint networks over the constraint language $\{=\}$ in $O( n \log n )$  partial queries.
	\end{lemma}
	
	\proof
	We consider the queries submitted to the user in line~\ref{qa:ask} of \qat and
	count how many times they can receive the answers \yes and \no. 
	
	For each \no answer in line~\ref{qa:ask} of \qat, 
	a new constraint will eventually be added to $L$. This new constraint $c$ 
	cannot be entailed by $L$ because the (complete) query generated in 
	line \ref{qa:gen-sol} of \qat must be accepted by $L$ and rejected by $c$. 
	In particular, $c$ cannot induce a cycle in $L$. It follows that at most 
	$n-1$ queries in line~\ref{qa:ask} are answered \no, each one entailing $O(\log n)$ more queries through the function \fscope and $O(1)$ through the function \fc.
	
	Now we bound the number of \yes answers in line~\ref{qa:ask} of \qat. Let $e_Y$ be an example generated by \qat in line~\ref{qa:gen-sol}. Let $B^{L\not\models}$ denote the set of constraints in $B$ that are not entailed by $L$.  In order to obtain a lower bound on the number of constraints in $B^{L\not\models}$ that $e_Y$ violates, we consider an assignment $\phi$ to $X$ that maps each connected component of $L$ to a value in $D$ drawn uniformly at random. We will show that the expected number of constraints that $\phi$ violates is $|B^{L\not\models}|/2$. Since \qat\ selects the assignment that \textit{maximizes} the number of violated constraints, it will follow that $e_Y$ violates at least half of $B^{L\not\models}$.
	
	By construction, the random assignment $\phi$ is accepted by  $L$. 
	Furthermore, each constraint $c$ in $B^{L\not\models}$ involves two variables belonging to distinct 
	connected components of $L$ so the probability that $\phi$ satisfies $c$ 
	is $|rel(c)|/|D|^2 = 1/|D|$, where $|rel(c)|$ denotes the number of tuples in $|D|^2$ that 
	belong to  the equality relation (the relation of the constraint $c$). 
	By linearity of expectation, the expected number of constraints that $\phi$ violates is therefore $|B^{L\not\models}| \cdot (1 - 1/|D|) \geq |B^{L\not\models}| \cdot 1/2$. As discussed in the previous paragraph, this implies in particular that $e_Y$ violates at least half the constraints in $B^{L\not\models}$. It follows that throughout its execution \qat will receive at most $\lceil \log |B| \rceil = \lceil \log n^2 \rceil$ yes answers at line~\ref{qa:ask}. 
	
	Putting everything together, the total number of queries that \qat\ may submit before it converges is bounded by $O(n \log n)$, as claimed.
	\qed
	
	\begin{lemma}
		\label{lem:ub-equalsup}
		If $|D| = 2$, then \qat learns constraint networks over the constraint language $\{=,\neq,>\}$ in $O( n \log n )$ partial queries.
	\end{lemma}
	
	\proof
	The proof follows the same strategy as that of Lemma~\ref{lem:ub-equal}, although the details are a little more involved. Again, we will count how many queries can be submitted to the user in line~\ref{qa:ask} of \qat.
	
	Each (complete) query submitted in line~\ref{qa:ask} that receives a negative answer will eventually add a new, non-redundant constraint to $L$. Observe that if $(L_{=}, L_{\neq},L_{>})$ denotes the partition of $L$ into sub-networks containing only constraints $=$, $\neq$ and $>$ respectively, then neither $L_{=}$ nor $L_{\neq}$ may contain a cycle; if $L_{>}$ does then the solution set of $L$ is empty and \qat will halt at line~\ref{qa:converge} the next time it goes through the main loop. Therefore, at most $3n$ queries may receive a negative answer in line~\ref{qa:ask}, each entailing $O(\log n)$ additional queries through the function \fscope and $O(1)$ through the function \fc.
	
	In order to bound the number of \yes answers in line~\ref{qa:ask} of \qat, consider an example $e_Y$ generated by \qat at line \ref{qa:gen-sol}. Let $B^{L\not\models}$ denote the set of constraints in $B$ that are not entailed by $L$.  Again, we claim that $e_Y$ violates at least half the constraints in $B^{L\not\models}$.
	
	We assume without loss of generality that $D = \{0,1\}$, interpreted as the Boolean values \true\ and \false.  Let ${\cal S}$ denote the set of connected components in the constraint network $L_{=,\neq}$ (the restriction of $L$ to constraints that are either equalities or disequalities). We say that a connected component $S \in {\cal S}$ is \emph{free} if there does not exist a constraint in $L$ of the form $X_i > X_j$ with either $X_i$ or $X_j$ in $S$. Because $L$ is satisfiable, free connected components $S$ have exactly two satisfying assignments $s,\overline{s}$, where $\overline{s}$ is the logical negation of $s$. All other components have exactly one satisfying assignment $s$.
	
	We construct a random assignment $\phi$ to $X$ as follows. For each connected component $S \in \cal{S}$, the restriction of $\phi$ to $S$ is either $s$ or $\overline{s}$ (chosen uniformly at random) if $S$ is free, and $s$ otherwise. By construction $\phi$ is accepted by  $L$, and for each variable 
	$X_k \in X$ that belongs to a free component, the probability that $\phi$ assigns $X_k$ to $1$ 
	is exactly $1/2$. It follows that, for each constraint $c$ in $B^{L\not\models}$, the probability 
	that $\phi$ violates $c$ is either $1/2$ (if $c$ is an equality or disequality, or a constraint $X_i > X_j$ involving exactly one free component) or $3/4$ (if $c$ is a constraint $X_i > X_j$ involving two free components). Overall, the expected number of constraints in $B^{L\not\models}$ that $\phi$ violates is at least $1/2 \cdot |B^{L\not\models}|$. In particular, there exists an assignment that violates at least half the constraints in $B^{L\not\models}$, and by the way \qat generates examples in line~\ref{qa:gen-sol}, $e_Y$ does as well.
	
	In conclusion, \qat will receive $\lceil \log |B| \rceil = O(\log n)$ \yes answers and $O(n)$ \no answers at line \ref{qa:ask}, plus $O(n \log n)$ answers within \fscope and \fc. The total number of queries made by \qat is therefore bounded by $O(n \log n)$.
	\qed
	
	\begin{lemma}
		\label{lem:algoless}
		If $|D| = 2$, then constraint networks on the language $\{ > \}$ can be learned in $O(n)$ partial queries.
	\end{lemma}
	
	\proof
	Suppose that the constraint network we are trying to learn has at least one solution. Observe that in order to describe such a problem, 
	the variables can be partitioned 
	into three sets:
	one for variables that must take the value $1$ (i.e., on the left side of a $>$ constraint),
	a second for variables that must take the value $0$ (i.e., on the right side of a $>$ constraint),
	and the third for unconstrained variables. In the first phase, we greedily partition variables into three sets, ${\cal L}, {\cal R}, {\cal U}$ initially empty and standing respectively for
	\emph{Left}, \emph{Right} and \emph{Unknown}.
	During this phase, we have three invariants:
	\begin{enumerate}  \itemsep=0pt
		\item There is no $X_i,X_j \in {\cal U}$ such that $X_i>X_j$ belongs to the target network
		\item $X_i \in {\cal L}$ iff there exists $X_j \in {\cal U}$ and a constraint $X_i>X_j$ in the target network
		\item $X_i \in {\cal R}$ iff there exists $X_j \in {\cal U}$ and a constraint $X_j>X_i$ in the target network
	\end{enumerate}
	
	We go through all variables of the problem, one at a time. 
	Let $X_i$ be the last variable picked. We query the user with an assignment where 
	$X_i$, as well as
	all variables
	in ${\cal U}$ are set to $1$, and all variables in ${\cal R}$ are set to $0$ (variables in ${\cal L}$ are left unassigned).
	If the answer is \yes, then there are no constraints between $X_i$ and any variable in ${\cal U}$, hence
	we add $X_i$ to ${\cal U}$ without breaking any invariant.
	Otherwise we know that $X_i$ is either involved in a constraint $X_j>X_i$ with $X_j \in {\cal U}$, or a constraint $X_i>X_j$
	with $X_j \in {\cal U}$. In order to decide which way is correct,
	we make a second query, where the value of $X_i$
	is flipped to $0$ and all other variables are left unchanged.
	If this second query receives a \yes answer, then the former hypothesis is true and we add $X_i$ to ${\cal R}$, otherwise, we add it to ${\cal L}$. Here again, the invariants still hold.
	
	At the end of the first phase, we therefore know that variables in ${\cal U}$ 
	have no constraints between them. However, they might be involved in constraints with variables in
	${\cal L}$ or in ${\cal R}$.
	In the second phase, we go over each variable $X_i \in {\cal U}$, and query the user with an assignment 
	where all variables in ${\cal L}$ are set to $1$, all variables in ${\cal R}$ are set to $0$ and $X_i$ is set to $1$.
	If the answer is \no, we conclude that there is a constraint $X_j>X_i$ with $X_j \in {\cal L}$ and therefore
	$X_i$ is added to ${\cal R}$ (and removed from ${\cal U}$). Otherwise, we ask the same query, but with the value of $X_i$ flipped to $0$. If the
	answer is \no, there must exist $X_j \in {\cal R}$ such that $X_i>X_j$ belongs to the network, hence $X_i$ is added to ${\cal L}$ (and removed from ${\cal U}$).
	Last, if both queries get the answer \yes, we conclude that $X_i$ is not constrained.
	When every variable has been examined in this way, variables remaining in ${\cal U}$ are not constrained.
	
	Once ${\cal L}, {\cal R}, {\cal U}$ are computed, we construct an arbitrary constraint network $C$ over $\{ > \}$ that is consistent with these sets. At this point, either $C$ is equivalent to the target network or our only assumption (the target network has at least one solution) was incorrect. We resolve this last possibility by submitting an arbitrary solution to $C$ to the user. If the answer is \yes, then we return $C$. Otherwise, the target network has no solution and we return an arbitrary unsatisfiable network over $\{ > \}$.
	\qed\\
	
	We are now ready to prove Theorem~\ref{thm:languages}.\\
	
	\proof[of Theorem~\ref{thm:languages}]
	We first consider the case $|D| = 2$. By Lemma~\ref{lem:ub-equalsup}, \qat learns constraint networks over any language $\Gamma \subseteq \{=,\neq,>\}$ in $O(n \log n)$ queries. Furthermore, if $\Gamma$ contains either $\{=\}$ or $\{\neq\}$ then this bound is optimal by Lemma~\ref{lem:lb-equal} and Lemma~\ref{lem:lb-neq}. This leaves the case of $\Gamma = \{>\}$. By Lemma~\ref{lem:algoless}, this language is learnable in $O(n)$ queries; this upper bound is tight since there are $\Omega(2^{n/2})$ non-equivalent constraint networks over $\{>\}$ on $n$ variables. (Take, for instance, the $2^{n/2}$ sub-networks of $C = \{ (X_i > X_{n/2+i}) \mid 1 \leq i \leq n/2 \}$ for $n$ even.) On the other hand, such constraint networks can have $\Omega(n)$ non-redundant constraints and \qat\ learns $O(1)$ constraints per call to \fscope.  Each of these calls to \fscope\ takes $\Omega(\log n)$ queries, so in the worst case \qat\ requires $\Omega(n \log n)$ queries. Combining this observation with Lemma~\ref{lem:ub-equalsup} we obtain that \qat\ learns networks over $\{ > \}$ (with domain size $2$) in $\Theta(n \log n)$ queries in the worst case.
	
	Now, assume that $|D| > 2$. If $\Gamma = \{=\}$ then by Lemma~\ref{lem:lb-equal} and Lemma~\ref{lem:ub-equal}, \qat learns networks over $\Gamma$ in $\Theta(n \log n)$ queries in the worst case and this bound is optimal. For every other language, Lemma~\ref{lem:rectangularity} establishes a universal worst-case lower bound of $\Omega(n^2)$ queries. A straightforward learning algorithm that examines all possible ordered pairs of variables and uses partial queries to determine the constraints of the target network on each pair will converge after $O(n^2)$ partial queries. Such constraints networks can have $\Omega(n^2)$ non-redundant constraints, so in the worst case \qat\ submits $\Omega(n^2 \log n)$ queries. This matches the general upper bound from Theorem~\ref{theo:complexity} since the basis has size $O(n^2)$.
	\qed


	\newcommand{\bdeg}{\texttt{bdeg}\xspace}

	\newcommand{\random}{{\tt Random}\xspace}
	\newcommand{\sudoku}{{\tt Sudoku}\xspace}
	\newcommand{\jigsaw}{{\tt Jigsaw}\xspace}
	\newcommand{\latin}{{\tt Latin}\xspace}
	\newcommand{\zebra}{{\tt Zebra}\xspace}
	\newcommand{\purdey}{{\tt Purdey}\xspace}
	\newcommand{\rlfap}{{\tt RLFAP}\xspace}
	\newcommand{\golomb}{{\tt Golomb}\xspace}
	
	\newcommand{\bk}{\mathcal{K}}
	
	\newcommand{\CR}{ ${\tt C\%}$}
	\newcommand{\Conv}{${\tt \#C}$}
	
	\newcommand{\AR}{ ${\tt A\%}$}
	
	\newcommand{\QR}{ ${\tt Q\%}$}
	
	\newcommand{\tmax}{{\tt t_{max}}}

	\newcommand{\QA}{${\tt \#Q_A}$\xspace}
	\newcommand{\QC}{${\tt \#Q_C}$\xspace}
	\newcommand{\QS}{${\tt\overline{|Q|}}$\xspace}
	\newcommand{\TA}{${\tt time_{A}}$\xspace}
	\newcommand{\TC}{${\tt time_{C}}$\xspace}
	\newcommand{\TIME}{${\tt \overline{t}}$\xspace}
	\newcommand{\TMAX}{${\tt t_{max}}$\xspace}

	\newcommand{\TO}{{$\tt TO$}\xspace}

	\section{Experimental Evaluation}
	\label{sec:exp}

	\newcommand{\RQ}[1]{\textbf{[Q#1}]}

	In this section, we experimentally evaluate \qat. 
	The purpose of our evaluation is to answer the following  questions:
	
	\begin{description}
		\item[\RQ{1}] {How does \qat behave in its basic setting? }
		\item[\RQ{2}] {How to make \qat faster to generate queries? }
		\item[\RQ{3}] {How effective is \qat when a background knowledge is provided?}
	\end{description}

	In the following subsections, we first describe the benchmark instances. 
	Second, we evaluate \qat in its basic setting. This
	baseline version allows us to observe that \qat may be subject to long 
	query-generation times.
	We then propose a strategy to make \qat faster in generating queries. 
	We validate this strategy on our benchmark problems. 
	Finally we evaluate the efficiency of \qat when a background knowledge 
	is provided. 
	This last experiment shows us that the number of queries required by \qat 
	to converge can dramatically decrease when the user is able to provide 
	some background knowledge about the problem to acquire. 
	
	For each of our experiments, \qat was run ten times on each problem 
	and the reported results are the averages of the ten runs. 
	For each run, we have set a time limit of one hour on the time to generate a query, after which 
	a time out (\TO) was reported. 
	All the results reported in this section were obtained with the version of \fc 
	that uses line \ref{qa:det:select-ex-v2} described in Section \ref{sec:qa:theo}. 
	We also tried the basic version that uses line \ref{qa:det:select-ex} described in 
	Algorithm \ref{alg:detect}. The results did not make 
	any significant difference. 
	All experiments were conducted using C++ platform\footnote{gite.lirmm.fr/constraint-acquisition-team/quacq-cpp} 
	on an Intel(R) Xeon(R) E5-2667 CPU, 2.9 GHz with 8 Gb of RAM. 
	
	The  performance of \qat is measured  according to  the following criteria: 
	
	\begin{itemize}
		\item[{[$|T|$]}]  size (i.e., number of constraints)
		of the target network $T$,
		\item[{[$|L|$]}] size of the learned network $L$, 
		\item[{[\QA{}]}]  total number of queries to learn a network $L$ equivalent to $T$, 
		\item[{[\QC{}]}] total number of queries to converge (i.e., until it is proved 
		that $L$ is equivalent to $T$),
		\item[{[\QS{}]}]  average size  of all queries, 
		\item[{[\TA{}]}] cumulated waiting time until a network ${L}$ equivalent to $T$ is learned,
		that is, time needed to generate all the queries until this  network ${L}$ equivalent to $T$ is found,
		\item[{[\TC{}]}] cumulated waiting  time  until convergence is reported,
		\item[{[\TIME{}]}] average time  needed to compute a query, 
		\item[{[\TMAX{}]}]  maximum waiting time between two queries, and
		\item[{[\Conv]}] number of runs that finished without triggering the 1-hour cutoff. 
	\end{itemize}

	%
	
	\subsection{Benchmark Problems}

	We evaluated \qat on a variety of benchmark problems whose characteristics are the following.

	\paragraph{Problem \purdey \cite{dell-magazine}.}
	Four families stopped by Purdey’s general store, 
	each to buy a different item. They all paid with different means. 
	Under a set of additional constraints
	given in the description, the problem is to match each family 
	with the item they bought and how 
	they paid for it. 
	This problem has a single solution.
	The target network of Purdey
	has 12 variables with domains of size 4 and 27 binary constraints.
	There are  three types  of variables, \emph{family},
	\emph{bought} and \emph{paid}, each of them containing four variables. 
	We initialized \qat with a basis of constraints of size 396 from  
	the language $\Gamma = \{\geq, \leq, <,>,\neq, = \}$.

	\paragraph{Problem \zebra.}
	The target network of the well-known Lewis Carroll's zebra problem is 
	formulated using 25 variables of domain size of 5
	with 5 cliques of $\neq$ constraints and 14 additional constraints 
	given in the description of the problem.
	The problem has a single solution.   
	We initialized \qat with a basis of 2700
	unary and binary constraints from the language
	$\Gamma =\{\geq, \leq, <,>,\neq, =, \circ_{val} , \shortparallel_{1}, \nshortparallel_{1}\}$,
	where $\circ_{val}$ denotes the unary relation $(x\circ val)$ with 
	$\circ \in \{\geq, \leq, <,>,\neq, =\}$ and $val\in 1..5$, and where 
	$ \shortparallel_{1}$ and $ \nshortparallel_{1}$
	respectively denote the
	distance relations $|x-y|=1$ and $|x-y|\neq 1$.

	\paragraph{Problem \golomb  \cite[prob006]{csplib}.}
	A Golomb ruler problem is to put a set of $n$ marks  on a ruler so that the
	distances between marks are all distinct. 
	This is encoded as a target network with $n$ variables corresponding to
	the $n$ marks, and constraints of varying arity. 
	We learned the target network of 350 constraints encoding the 8-marks ruler.
	We initialized \qat with a basis of 1680  binary, ternary and quaternary constraints 
	from the language $\Gamma = \{\geq, \leq, <,>,\neq, =, \shortparallel^{zt}_{xy}, \nshortparallel^{zt}_{xy} \}$, 
	where $\shortparallel^{zt}_{xy}$ and $\nshortparallel^{zt}_{xy}$ 
	respectively denote the distance relations $|x - y|=|z - t|$ and $|x - y|\neq|z - t|$. 
	Observe that when $x$ and $z$,  or $y$ and $t$ represent the same variable, 
	$\shortparallel^{zt}_{xy}$ and $\nshortparallel^{zt}_{xy}$ yield ternary constraints.

	\paragraph{Problem \random. }
	We generated a binary random target network 
	with 50 variables, domains of size 10, and  
	122 binary constraints. 
	The 122 binary constraints are iteratively and randomly selected from the complete 
	graph of binary constraints from the language $\Gamma=\{\geq, \leq, <,>,\neq, =\}$. 
	When a constraint is randomly selected it is inserted in 
	the target network only if this pair of variables is not already linked 
	by a constraint and if the new constraint is not implied by the already selected 
	constraints.  
	\qat is initialized with a basis of constraints containing the complete 
	graph of 7350 binary constraints from~$\Gamma$.

	\paragraph{Problem \rlfap.}
	The Radio Link Frequency Assignment Problem  is to provide communication channels from 
	limited spectral resources so as to avoid interferences between channels \cite{CabonGLSW99}.
	The constraint network
	of the instance we selected has 50 variables
	with domains of size 40 and 125 binary constraints (arithmetic and distance constraints).
	We initialized \qat with a basis of 12,250 constraints from 
	the language $\Gamma = \{\geq, \leq, <,>,\neq, =, =^{val}_{xy}, >^{val}_{xy} \}$, 
	where $=^{val}_{xy}$ and $>^{val}_{xy}$ respectively denote the distance relations $|x-y|=val$ and $|x-y|>val$,
	and $val\in\{12,14,28,35,56,84,238\}$.

	\paragraph{Problem \sudoku.}
	The Sudoku logic puzzle is a $9\times 9$ grid.
	It must be filled
	in such a way that all the rows, all the columns and
	the 9 non-overlapping $3\times 3$ squares contain the numbers 1 to 9. 
	The target network of Sudoku has  81 variables with domains of size 9 and
	810 binary  $\neq$ constraints on rows, columns and squares. 
	\qat is initialized with a basis $B$ of 19,440 binary constraints from the
	language $\Gamma = \{\geq, \leq, <,>,\neq, = \}$.

\begin{figure}[tb]
		\begin{center}
			\includegraphics[width=100pt]{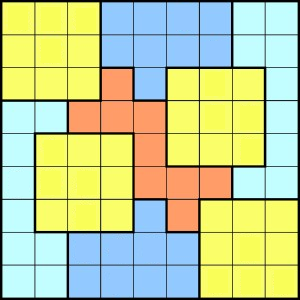}
		\end{center}
		
		\caption{Our  instance of \jigsaw problem.} \label{fig:jigsaw}
	\end{figure}

	\paragraph{Problem \jigsaw.}
	The Jigsaw Sudoku is  a variant of Sudoku in which the $3\times 3$ squares  
	are replaced by  irregular shapes. We used the instance of Jigsaw Sudoku 
	displayed in Figure \ref{fig:jigsaw}. 
	The target network has  81 variables with domains of size 9 and 811 binary  $\neq$
	constraints on rows, columns and shapes. 
	\qat is initialized with a basis $B$ of 19,440 binary constraints from the
	language $\Gamma = \{\geq, \leq, <,>,\neq, = \}$.
	
	%

	\subsection{\RQ{1} \qat in its basic setting}\label{sec:exp:basic}
	
	When \qat is used in its basic setting, we denote it by \qatb.
	What we call the basic setting is when, in line \ref{qa:gen-sol} of Algorithm \ref{alg:qa}, 
	\qat uses the function \genexb described in Algorithm~\ref{alg:genex0}. 
	\genexb computes a complete assignment on $X$ satisfying the 
	constraints in $L$ and 
	violating at least one constraint from $B$. 
	We build a network $C$ that contains the constraints 
	from the network $L$ already learned (line \ref{alg:genex0:initL}), 
	plus  a reification of the constraints in $B$. 
	A Boolean $b_i$ is introduced for each  $c_i\in B$. This Boolean is forced to be true 
	if and only if the constraint $c_i$  is satisfied  (line~\ref{alg:genex0:reif}). 
	We then force the sum of $b_i$'s not to be equal to $|B|$ (line~\ref{alg:genex0:sum}). 
	Function $solve$ is called on $C$ (line~\ref{alg:genex0:solve}) and 
	returns a solution of $C$, or $\bot$ if no solution exists. 
	Finally, the projection  on $X$ of the solution is returned  (line~\ref{alg:genex0:return}). 
	The constraint solver inside $solve$ uses the \texttt{dom/wdeg} variable ordering heuristic \cite{bouetalECAI04} and a random value selection.

	\begin{algorithm}[tbp]
			$C \gets L$\;\label{alg:genex0:initL}
			\lForEach{$c_i\in B$}{
				$C\gets C\cup \{b_i\leftrightarrow c_i\}$\label{alg:genex0:reif}}	
			$C\gets C \cup \{\sum b_i \neq |B|\}$\;\label{alg:genex0:sum}
			$e\gets solve(C)$\; \label{alg:genex0:solve}    	
			\Return $e[X]$\;\label{alg:genex0:return}
		\caption{$\genexb(X,B,L)$}
		\label{alg:genex0}
	\end{algorithm}

	\begin{table}[tbp]
		\centering
		\caption{\qatb. All  results are averages of ten runs (time in seconds).
		} 
		\begin{tabular}{|l|rr|rrr|rrrr|r|}
			\hline
			{ Instance} &$|T|$ &$|L|$ & \QA &  \QC  &  \QS/$|X|$ & \TA & \TC& \TIME & \TMAX& \Conv\\ \hline 
			\hline
			\purdey & 
			27 & 26.2 & 175.3  & 177.1  & 5.0/12 & 0.08 & 0.09 & 0.00 & 0.01  & 10 \\\hline
			\zebra &
			64 & 61.1 & 555.6  & 555.8   & 8.1/25 & 2.54 & 2.54 & 0.00 & 1.50  & 10 \\ \hline
			\golomb &
			350 & 96.4 & 351.5 & 351.5  & 4.8/8  & 116.39 & 217.34 & 0.33 & 8.70 & 10  \\ \hline
			\random &
			122 & 122.0 & 1 082.2 & 1 092.0  & 20.8/50 & 2.08 & 85.94& 0.08 & 83.80 & 7 \\ \hline
			\rlfap &
			125 & 98.5 & 1 103.6 & --   & --    & 43.35  & --   & --     &  \TO   & 0       \\ \hline	
			\sudoku &
			810 & 775.7  & 6 849.9 & --  & --  & 214.16  &  --    & --    &  \TO   & 0       \\ \hline
			\jigsaw &
			811 & 764.0 & 6 749.6 & --   & --   & 224.18  & --   & --    &    \TO & 0       \\ \hline
			\multicolumn{11}{r}{\TO = 1 hour}
		\end{tabular}%
		\label{tab:Q0}%
	\end{table}%

	Table \ref{tab:Q0} reports the results of running \qatb on all our benchmark problems. 
	The first observation we can make by looking at the table is that there are 
	only four problems  on which \qatb has been able to converge in all 
	of the ten runs (\purdey, \zebra, \golomb) or in  some of them  (\random). 
	For \random, on which \qatb converges 7 times out of 10,  
	Table \ref{tab:Q0} reports the averages of these 7 runs. 
	
	We  first focus our attention on these  four problems: \purdey, \zebra, \golomb, and \random. 
	Let us first compare  the columns $|T|$ and $|L|$. On \purdey 
	and \zebra, we observe that the size of $L$ is slightly smaller than 
	the size of the target network $T$. This is due to a few 
	constraints that are redundant wrt to some subsets of $T$. 
	On \golomb, $|L|\ll|T|$ (96 and 350 respectively) because 
	our target network with  all quaternary constraints  $|X_i - X_j| \neq |X_k - X_l|$ 
	contains a lot of redundancies  \qatb detects convergence before learning them. 
	Finally, as  \random does not have 
	any structure, it does  not contain any  redundant constraint, and $|L|=|T|$. 
	The column  \QS/$|X|$ shows us that the queries asked by \qatb are often 
	much shorter than $|X|$. The average size \QS of queries varies from 
	one third to one half of $|X|$. 
	The  the number of queries \QC
	is two to seven times smaller than the size of the basis $B$. This 
	means that each positive query leads to the removal of several constraints from $B$. 
	Let us now compare the costs of finding the right network (columns \QA and \TA) 
	and the costs of converging (columns \QC and \TC).  
	This tells us a lot about the end of the learning process. 
	On  \purdey and \zebra, \QA and \TA are similar to \QC and \TC (respectively), which means 
	that \qatb learns constraints until the very end of the process. 
	On \golomb, \QA and \QC are again similar, but \TC is much larger than  \TA. 
	The reason is that after having learned all the constraints necessary to 
	have a $L$ equivalent to $T$, 
	\genexb spends 100 seconds to show that $L\models B$, which proves convergence. 
	On \random, we observe yet another behavior. 
	As on \golomb, \TC is  much larger than \TA (almost two orders of magnitude larger), but 
	\QC is also  larger than \QA. The reason is that  \qatb has found a network $L$ 
	equivalent to $T$ ten queries before the end  and  spends the end of the 
	learning process generating complete queries that are positive and that allow 
	\qatb  to remove useless constraints from  $B$ and finally prove convergence. 
	This last phenomenon is probably due to the sparseness of $T$ in \random. 
	%
	The columns \TIME and \TMAX tell us that  most queries are very easy to 
	generate (from milliseconds to one third of a second in average) and that 
	most  of the time is in fact 
	consumed by generating the last positive queries. 
	\random is  an extreme case where the very last query consumes forty times the time 
	needed for the whole process of learning $T$. 
	
	Let us now move our attention to the last three problems in Table \ref{tab:Q0}, 
	namely, \rlfap, \sudoku, 
	and \jigsaw.  
	On these  problems,  on each of the  ten runs, 
	\qatb  reaches the 1-hour cutoff on the time to generate a query. 
	However we see that \QA and \TA, which represent the cost of learning 
	a network equivalent to $T$ without having proved convergence, are reported in the table. 
	For all the runs and all problems, \qatb has found a network $L$  
	equivalent to  $T$ 
	before reaching the cutoff. 
	This is the proof of convergence that leads \qatb to the time-out. 
	\QA and \TA  represent the cost of learning a network equivalent to $T$ 
	but \qatb does not know it is the target. 
	Similarly to the first four problems, the number of queries required to
	learn a network equivalent to $T$ is significantly smaller than the size of 
	$B$ (from three to eleven times smaller). 
	
	From  this first experiment we  conclude that  \qatb 
	learns small constraint networks in a number of queries 
	always significantly smaller than the size of the basis 
	and generates queries in very short times. 
	However, as soon as the size of the target network increases, 
	the time to generate the last queries becomes prohibitive for 
	an interactive learning process.

	\subsection{\RQ{2} Faster query generation}\label{sec:fast-gen}

	\subsubsection{\genexc: Generating (partial) queries with a time limit}\label{sec:exp:genex1}

	The experiments in Section \ref{sec:exp:basic} have shown that 
	\qatb can be subject to excessive waiting time between two queries. 
	This prevents its use in an interactive process where a human is in the loop. 
	In this section we propose a new version of \genex that fixes this weakness. 
	We start from the observation that the example generated by \genex 
	in line \ref{qa:gen-sol} of Algorithm \ref{alg:qa} does not need to 
	be an assignment on $X$. 
	Any partial assignment is satisfactory as long as it 
	does not violate any constraint from $L$ and violates at least one constraint from $B$. 
	We propose thus  \genexc, a new version of \genex 
	that quickly returns a partial assignment on a subset $Y$ of $X$ accepted by  $L$ 
	and violating at least one constraint from $B$. The main idea 
	is to modify the function $solve$ so that it can be called with a cutoff. 
	
	Function $solve(C,S,obj,ub)$ takes as input a set $C$ of constraints to 
	satisfy, a set $S$ of variables that must be included in the assignment, 
	a parameter $obj$ to maximize, and an upper bound $ub$ on the time 
	allocated. 
	$solve$ returns a pair $(e_Y,t)$ where $e_Y$ is an assignment 
	on a set $Y$ of variables containing $S$, 
	and $t$ is the time consumed by $solve$. 
	If $solve$  proves that $C$ is inconsistent
	(that is, it found a set $Y$ containing $S$ for which every assignment on $Y$ 
	either violates $C[Y]$ or leads to arc inconsistency on $C$), 
	it returns the pair $(\bot,t)$ where $t$ is the time 
	needed to prove that $C$ is inconsistent. 
	If the allocated time $ub$ is not sufficient to find a satisfying assignment or prove an inconsistency, 
	$solve$ returns the pair $(nil, ub)$.
	Otherwise, $solve$ returns a pair $(e_Y,t)$ 
	where  $e_Y$ is an assignment accepted by $C$ 
	and with highest value of  $obj$ found during the allocated time $ub$, 
	and $t$ is the time consumed. 
	When $solve$  is called with $obj=nil$, there is nothing to maximize and the 
	first satisfying assignment (on $S$) is returned. 
	The function $solve$ uses the $\bdeg$  variable ordering heuristic \cite{TsourosSB19}.
	$\bdeg$ selects the variable involved in a maximum number of constraints from  $B$. 
	By following $\bdeg$, $solve$ tends to generate assignments  that violate more 
	constraints from  $B$, so that in case of \yes answer, 
	the size of $B$ decreases faster.

	\begin{algorithm}[tb]
			$\Time \gets 0$\;
			
			\ForEach{$c\in B$\label{alg:genex1:for}}{
				$(e_{var(c)},t)\gets solve(L\cup\{\neg c\},var(c),nil,+\infty)$\;    \label{alg:genex1:solve1}
				$\Time \gets \Time + t$\; \label{alg:genex1:time1}
				
				\uIf{$e_{var(c)} = \bot $}{	
						mark $c$ as redundant;	
						$L\gets L\cup \{c\}$; 
						$B\gets B\setminus \{c\}$\;\label{alg:genex1:implied1}
					}
				\Else{
					$(e'_Y,t')\gets solve(L\cup\{\neg c\},var(c), |Y|, \cutoff-\Time)$\; 		\label{alg:genex1:solve2}
					$\Time \gets \Time +t'$\;
					\lIf{$e'_Y = nil$}{\Return $e_{var(c)}$\label{alg:genex1:return1}}
					\uIf{$e'_Y = \bot$}{	
						mark $c$ as redundant;	
						$L\gets L\cup \{c\}$; 
						$B\gets B\setminus \{c\}$\; \label{alg:genex1:remove2}}
					\lElse{\Return $e'_Y$\label{alg:genex1:return2}}
			}}
			remove all constraints marked as  redundant from $L$\;\label{alg:genex1:redundant} 
			\Return $\bot$\;\label{alg:genex1:nil} 
		\caption{$\genexc(X,B,L)$
		}
		\label{alg:genex1}
	\end{algorithm}

	Algorithm \ref{alg:genex1} describes \genexc. \genexc takes as input 
	the set of variables $X$,
	a current basis of constraints $B$,
	a current learned network $L$
	and a timeout parameter $\cutoff$.
	\genexc iteratively picks a constraint $c$ from $B$ until a satisfying assignment 
	is returned or $B$ is exhausted (line \ref{alg:genex1:for}). 
	The call to $solve$ in line \ref{alg:genex1:solve1} computes an assignment $e_{var(c)}$ on ${var(c)}$
	violating $c$ and accepted by  $L$.
	The time $t$ needed to compute $e_{var(c)}$ is added to the time counter (line \ref{alg:genex1:time1}). 
	If $solve$ returns $e_{var(c)}=\bot$ (i.e., $L\cup\{\neg c\}$ is inconsistent), 
	$c$ is marked as redundant because it is implied by $L$. 
	$c$ is then removed from $B$ and added to $L$ (line~\ref{alg:genex1:implied1}). 
	Adding $c$ to $L$ is required to avoid that \qat will later try to learn this 
	constraint which is no longer in $B$. 
	If $solve$ returns an assignment $e_{var(c)}$ different from $\bot$, 
	\genexc enters a second phase during which 
	a second call to  $solve$ will use the remaining amount of time, $\cutoff-time$,  
	to compute an assignment $e'_Y$ violating $c$ and accepted by $L$, 
	whilst maximizing $|Y|$ (line \ref{alg:genex1:solve2}). 
	If  no such assignment is found in the remaining time, $solve$ returns an $e'_Y$ equal to $nil$ 
	and \genexc returns the $e_{var(c)}$ found by the first call to $solve$ (line \ref{alg:genex1:return1}). 
	If $solve$  proved the inconsistency of $L\cup\{\neg c\}$ over a 
	given scope $Y$ (i.e., $e'_Y=\bot$), 
	$c$ is marked as redundant, removed from $B$, and  added to $L$ (line \ref{alg:genex1:remove2}), 
	exactly like line~\ref{alg:genex1:implied1}. 
	\genexc then goes back to line \ref{alg:genex1:for} to  select a new constraint from $B$. 
	Otherwise (i.e., $e'_Y\notin \{nil,\bot\}$), 
	\genexc returns the assignment  $e'_Y$ with the largest  size of $Y$ 
	that has been found in the allocated time (line \ref{alg:genex1:return2}). 
	Finally, if all constraints in $B$ have been processed without finding 
	a suitable assignment (line \ref{alg:genex1:for}), 
	this means that all the constraints that were in $B$ were implied by $L$. 
	The learning process has thus converged. 
	We just need to remove all constraints marked as redundant 
	from $L$ (line \ref{alg:genex1:redundant}) 
	and \genexc  returns  $\bot$ (line \ref{alg:genex1:nil}). 
	It is not necessary to remove the redundant constraints but it usually 
	makes the learned network more compact and easier to understand.

	\subsubsection{Evaluation of \genexc}\label{sec:exp:cutoff}
	
	We made the same experiments as in Section \ref{sec:exp:basic} but instead 
	of using \qatb, we used \qatc, that is, \qat calling \genexc.  
	We have set the cutoff  to one second so that the acquisition process 
	remains comfortable in the case where the learner interacts with a human. 
	Table \ref{tab:Q1} reports the results for the same measures as 
	in Table \ref{tab:Q0}.
	
	\begin{table}[tbp]
		\centering
		\caption{\qatc with $\cutoff$ = 1 second. 
		}
		\begin{tabular}{|l|rr|rrr|rrrr|r|}
			\hline
			{Instance}		&  $|T|$ &  $|L|$ & \QA &  \QC  & \QS/$|X|$ & \TA & \TC & \TIME & \TMAX& \Conv\\ \hline \hline
			
			\purdey    & 27  & 26.3  & 172.3  & 179.1  & 5.3/12 & 0.09 & 0.20 & 0.00 & 0.03 &10 \\ \hline
			\zebra     & 64  & 61.2   & 556.5  & 561.2   & 8.1/25  & 3.92   & 4.27   & 0.01 & 2.31 &10\\ \hline
			\golomb    & 350 & 98.0   & 377.2  & 377.2    & 4.2/8  & 135.32 & 138.06 & 0.36 & 8.47 &10 \\ \hline
			\random    & 122 & 122.0 & 1 059.7 & 1 064.8  & 20.6/50 & 2.17   & 2.20   & 0.00 & 0.07 &10\\ \hline
			\rlfap     & 125 & 125.0 & 1 167.8 & 1 167.8  & 19.4/50 & 42.82  & 42.82  & 0.04 & 1.05 & 10 \\ \hline
			\sudoku    & 810 & 810.0 & 6 939.0 & 6 941.6  & 21.5/81 & 164.05 & 164.32 & 0.02 & 1.01 & 10 \\ \hline
			\jigsaw    & 811 & 811.0  & 6 874.5  & 6 880.2 & 21.0/81 & 225.40 & 231.04 & 0.03 & 1.01 &10 \\ \hline
		\end{tabular}%
		\label{tab:Q1}%
	\end{table}
	
	The main information that we extract from Table \ref{tab:Q1} is that the use 
	of \genexc has a dramatic impact on the time consumption of generating queries. 
	The cumulated generation time for all queries until convergence never exceeds five minutes 
	on any run on any problem 
	whereas \qatb was reaching the one-hour cutoff for a query on all runs
	on three of the problems. 
	Even on the problems where \qatb was converging, \qatc can show 
	a significant speed up. 
	For instance,  on  \random,  \qatb was taking a long time to prove that the learned 
	network was equivalent to the target one (\TC $\gg$ \TA). 
	With \qatc, \TC and \TA are almost equal and are close to the value of \TA of \qatb.

	We could have expected  that the generation of shorter queries at the end of 
	the learning process leads to an increase in the overall  number of queries 
	for \qatc because  shorter positive queries lead to fewer redundant constraints detected. 
	But, when comparing \QA and \QC in Tables \ref{tab:Q0} and \ref{tab:Q1}, 
	we see  that  the increase is negligible. There is a -2\% to +2\% difference 
	on most problems. 
	The exceptions are \golomb and \rlfap, on which \qatc exhibits an increase 
	of 6\% and 7\% respectively. 
	Similarly, 
	\QS/$|X|$ is essentially the same for \qatb and \qatc. 
	As a last observation on Table \ref{tab:Q1}, 
	it can seem surprising that \TMAX is more than eight seconds on \golomb despite 
	the cutoff of one second in \genexc.  This is because  during the learning process, 
	\genexc repetitively finds redundant constraints  without asking a 
	question to the user (line \ref{alg:genex1:implied1} in Algorithm~\ref{alg:genex1}). 
	
	This experiment shows us that the introduction of a cutoff in the generation of 
	examples completely solves the issue of extremely long waiting times at the end 
	of the process. 
	\qatc learned all our benchmarks problems with extremely fast query generation. 
	The only price to pay is a slight increase in number of queries in two of the 
	problems.

	\subsection{\RQ{3} \qat with background knowledge}\label{sec:exp:bk}
	
	In practical applications, it is often the case that the user already knows some 
	of the constraints of her problem. These constraints can be known because they 
	are easy  to express, or because they are implied by the structure of the problem, 
	or because they have been learned by another tool. 
	For instance, given a solution to a sudoku or to a jigsaw sudoku, 
	\modelseeker would be able to learn that all the cells in a row must 
	take different values. 
	We can also inherit constraints from a past/obsolete model that needs 
	to be updated because some changes have occurred in the problem. 
	For instance, if new workers 
	have joined the company, we need to learn constraints on them, 
	but the rest of the problem remains unchanged. 
	This set of already known constraints will be called the \emph{background knowledge}. 
	
	\qat can easily be adapted to handle the case of a background knowledge. 
	In the following, a background knowledge will be a set $K$ of constraints, 
	where $K$ is the part of the target problem that we already know, that is, $K\subseteq T$.
	Instead of calling \qat with an empty network $L$ and a basis $B$, 
	\qat is called with $L$ initialized to $K$ and the basis initialized 
	to $B\setminus \{K\cup \bar{K}\}$, where  $\bar{K}=\{c\mid \neg c \in K\}$.

	We performed a first experiment with \purdey, \zebra, \golomb, and \jigsaw. 
	In \purdey, we assume that the user was able to express that 
	if  four different families buy four different items  with four different paying means,
	then 
	there is a clique of dis-equalities on the four variables representing families, 
	a clique of dis-equalities on the  variables representing items to buy, 
	and also a clique on the variables representing  paying means. 
	Similarly, in \zebra, we assume that the user was  able to express that if 
	there are five people of five different nationalities, 
	there is a clique of dis-equalities on the five variables representing nationalities. 
	Idem on colors of houses, drinks, cigarettes and pets. 
	In \golomb, we assume the user was able to express the symmetry breaking constraint 
	$X_i<X_{j}$ for all pairs of marks $i,j, i<j$. 
	Finally, in  \jigsaw, we  assume that the user ran 
	\modelseeker on the solutions of a few instances of these problems and learned 
	that there is a clique of dis-equalities on all the rows and all the columns. 
	
	\begin{table}[tbp]
		\centering
		\caption{\qatc when the cliques of disequalities are already given in \purdey, \zebra, 
			and \jigsaw, and  the symmetry-breaking constraints  in  \golomb.}
		\begin{tabular}{|l|rrr|rrr|rrrr|}
			\hline
			{Instance}	&  $|T|$	&  $ |K|$ &  $ |L|\setminus |K|$ & \QA &  \QC & \QS & \TA & \TC & \TIME & \TMAX\\ \hline \hline
			
			\small  \purdey  	
			& 27 & 18  & 9.0 & 70.8 & 81.5 & 5.6/12 & 0.05 & 0.24 & 0.00 & 0.03 	\\ \hline 
			\small   \zebra 	
			& 64 & 50  & 14.0    & 185.6    & 199.2   & 7.1/25  & 3.91   & 4.87   & 0.02 & 2.92 	\\ \hline 
			\small   \golomb
			& 350 & 28 & 70.0    & 246.5    & 246.5    & 5.1/8  & 132.10 & 139.51 & 0.53 & 6.00     \\ \hline 
			\small   \jigsaw
			& 811 & 648 & 163.0   & 1 688.8 & 1 715.0  & 20.5/81 & 175.37 & 201.29 & 0.12 & 1.03  \\ \hline 
		\end{tabular}%
		\label{table:bk:cliques}
	\end{table}%
	
	Table \ref{table:bk:cliques} reports the results when running \qatc with an 
	$L$ initialized to $K$ as described above for the four problems. 
	The main observation is that the number of queries asked by \qatc 
	significantly decreases. 
	This decrease in number of queries goes from a factor 
	1.5 for \golomb, where $|K|/|T|\approx 0.08$, 
	2.2 for \purdey, where $|K|/|T|\approx 0.67$, 
	2.8 for \zebra, where $|K|/|T|\approx 0.78$, 
	to 4.0 for \jigsaw, where $|K|/|T|\approx 0.80$.  
	This shows  that 
	the larger the number of constraints already known, 
	the greater the decrease in number of queries. 
	These are good news because the number of queries is a critical criterion 
	when the user is a human. 
	The second interesting information we learn from this experiment is that 
	most of the other characteristics of \qatc are essentially the same whatever 
	\qatc is provided with an initial background knowledge or not.  
	The only exception is the average  time to generate a  query, \TIME, 
	that tends to increase in the presence of a background knowledge. This is 
	not surprising because  we know that this is 
	when we are close to the end of the acquisition process ---when $L$ is large--- 
	that  query generation  costs the most. 
	But this increase only occurs because our queries are very fast to generate, 
	faster than the cutoff of one second. If queries were becoming too long to generate, 
	the cutoff would force shorter queries.

	We 
	performed a second experiment on \random, \rlfap, \sudoku, and  \jigsaw, that 
	are  the problems on which  \qatc asks the more queries. 
	Similarly to the previous experiment, we called  \qatc with a 
	learned network $L$ partially filled with a background knowledge $K$ 
	and a basis initialized to $B\setminus \{K\cup \bar{K}\}$. 
	We varied the size of $K$ by randomly picking from 0\% to 90\% of the 
	constraints in the target network. 
	
	Figure \ref{fig:queriesbk} reports the ratio \QC-w$K$/\QC-wo$K$ 
	of the number of queries that \qatc requires to converge with a given $K$ 
	on the number of queries required to converge without any $K$. 
	These results show that when the size of $K$ increases, the number of 
	queries decreases. 
	On \rlfap, \qatc drops from 1168 queries 
	without $K$ to only 13 queries  when $K$ contains 90\% of the target network. 
	Importantly, on all problems the decrease is strongly correlated to the amount of 
	background knowledge provided (slope almost linear). 
	This is very good news because this means that it always deserves 
	to add more background knowledge. 
	%
				%
					%
					%
					%
					%
					%

	\begin{figure}
		\begin{center}
			\begin{tikzpicture}
				
				\begin{axis}[width=12cm,height=6cm,
					xlabel=$|K|/|T|$,
					ylabel=\QC-w$K$/\QC-wo$K$
					]
					\addplot plot[smooth,mark=square,mark size=2pt] coordinates {
						(0  ,  1                 )
						(0.1 , 0.847577009767092 )
						(0.2 , 0.724267468069121 )
						(0.3 , 0.608940646130729 )
						(0.4 , 0.505634861006762 )
						(0.5 , 0.405240420736289 )
						(0.6 , 0.309072126220887 )
						(0.7 , 0.225957926371149 )
						(0.8 , 0.137396694214876 )
						(0.9 , 0.066303531179564 )
					};
					
					\addplot plot[smooth,mark=*,mark size=2pt] coordinates {
						(0,     1                 )
						(0.1 ,  0.842010618256551 )
						(0.2 ,  0.709111149169378 )
						(0.3 ,  0.556430895701319 )
						(0.4 ,  0.43235143003939  )
						(0.5 ,  0.305617400239767 )
						(0.6 ,  0.217845521493406 )
						(0.7 ,  0.123822572358281 )
						(0.8 ,  0.033481760575441 )
						(0.9 ,  0.011303305360507) 
					};
					
					\addplot plot[smooth,mark=x,mark size=5pt] coordinates {
						(0     , 1                  )
						(0.1  ,  0.854644462371788  )
						(0.2  ,  0.717024893396335  )
						(0.3  ,  0.596966117321655  )
						(0.4  ,  0.491745418923591  )
						(0.5  ,  0.388080557796473  )
						(0.6  ,  0.30093926472283   )
						(0.7  ,  0.213466635934079  )
						(0.8 ,   0.132577503745534  )
						(0.9,    0.06059121816296   )
					};
					
					\addplot plot[smooth,mark=+,mark size=5pt] coordinates {
						(0   ,   1)
						(0.1 ,   0.85817892298784)
						(0.2 ,   0.724392009264621)
						(0.3 ,   0.611334684423856)
						(0.4 ,   0.496598147075854)
						(0.5 ,   0.40036189924725)
						(0.6 ,   0.305153445280834)
						(0.7 ,   0.21645917776491)
						(0.8 ,   0.138578459756804)
						(0.9 ,   0.065040532715692)
					};
					
					\legend{\random\\ \rlfap\\ \sudoku\\ \jigsaw\\}
					
				\end{axis}
			\end{tikzpicture}
		\end{center}
		\caption{\QC-w$K$/\QC-wo$K$ for \qatc when varying the size of $K$.} \label{fig:queriesbk}
	\end{figure}
	
	We do not report any result on \qatb with background knowledge. 
	Whatever the amount of background knowledge provided, 
	\qatb suffers from the same drawback as \qatb without background 
	knowledge: the last few queries are prohibitively expensive to generate. 
	On \rlfap, \sudoku, and  \jigsaw, \qatb cannot converge on any run of any size of $K$ 
	within  the one-hour time limit  on query generation time. 

	\subsection{Discussion}
	
	These experiments tell us several important features of \qat. 
	These experiments show us that \qat can learn any kind of network, whatever 
	all their constraints are organized with a  specific structure (such as \sudoku), 
	some of their constraints have a structure (\purdey, \zebra, \golomb, \jigsaw), 
	or they  have no structure at all (\random,  \rlfap). 
	
	A second  general  observation is that \qat learns a network in a number of queries 
	always significantly smaller than the size of the basis. This confirms 
	that \qat is able to select  the queries  in a way that makes them very 
	informative for the learning process. 
	However, the experiment in Section \ref{sec:exp:basic} shows that 
	when \qat is used in its basic version, the time to generate queries can be prohibitive, 
	especially when interacting with a human. 
	We indeed observe that when we are close to  the end of the learning process, 
	it can be extremely time consuming to generate a  complete example at the start of each 
	loop of acquisition of  \qat. 
	
	The experiment in Section \ref{sec:exp:cutoff}  shows 
	that a simple adaptation of the way examples are generated
	(see function \genex.1 in Section \ref{sec:exp:genex1}), 
	allow us to monitor the CPU time needed to generate an example with a cutoff. 
	In the experiment we see that a cutoff of one second leads to a very smooth 
	interaction between the learner and the user. 
	It is important to bear in mind that even with a cutoff, \qat 
	ensures the property of convergence. 
	
	All experiments in Sections \ref{sec:exp:basic} and \ref{sec:exp:cutoff} were 
	performed in the scenario where we do not have any background knowledge about the problem. 
	In these experiments, \qat is always initialized with an empty learned network $L$. 
	As a result, the number of queries necessary on some of the benchmark problems 
	can seem unrealistic for a real use, especially in the presence of a human user. 
	But it is often the case that the user is able  to 
	express  some of the constraints of the problem, 
	or that an initial subset 
	of the constraints can be learned 
	with a tool such as \modelseeker or \arnold. 
	Such  tools are able to learn constraints with very few examples when 
	these constraints follow some specific   structure. 
	In such scenarios, \qat can be used as a complement that will learn the missing 
	constraints, that is, the constraint that cannot be captured in a specific 
	structure recognized by these tools. 
	Our experiment in Section \ref{sec:exp:bk} shows that \qat is extremely good 
	when it is provided with a background knowledge in the form of a set of constraints. 
	The larger the set of constraints given as a background knowledge, 
	the fewer queries needed to learn the  network.

	
	\section{Conclusion}
	\label{sec:conclusion}
	
	We have proposed \qat, an algorithm that learns constraint 
	networks by asking the user to classify partial assignments
	as positive or negative. 
	Each time it receives a negative example,
	the algorithm converges on a constraint of the target network
	in a logarithmic number of queries. 
	Asking the user to classify partial assignments allows \qat to  converge on
	the target constraint network in a polynomial number of
	queries (as opposed to the exponential number of queries required when 
	learning with membership queries only). 
	We have shown that \qat is optimal on certain constraint languages and 
	that it is close to optimal (up to $\log n$ worse) on  others.
	Furthermore, as opposed to other 
	techniques, the user does not need to provide 
	positive examples to learn the target network. 
	This  can be very useful when
	the problem has never been solved before. 
	Our experiments show that \qat in its basic version can be time consuming 
	but they also  show that \qat can be parameterized with a 
	cutoff on the waiting time that allows it to generate queries quickly. 
	These experiments also show that \qat  can learn any kind of network, 
	whatever its constraints follow  a specific structure (such as matrices) or not. 
	More, we  observed that \qat behaves very well in the presence 
	of a background knowledge. 
	The larger the background knowledge, the fewer the queries required to 
	converge on the target network.  
	This last feature makes \qat a perfect candidate to learn missing 
	constraints in a partially filled constraint model. 
	As a last comment, we should bear in mind that all the improvements of \qao 
	already published in the literature (for instance 
	\cite{TsourosSS18,BessiereCDLMB14,daoetalIJCAI16,AddiBEL18,TsourosSB19}), can be used 
	with \qat.

\bibliographystyle{plain}

\begin{thebibliography}{10}
	
	\bibitem{AddiBEL18}
	H.A. Addi, C.~Bessiere, R.~Ezzahir, and N.~Lazaar.
	\newblock Time-bounded query generator for constraint acquisition.
	\newblock In {\em Proceedings of the 15th International Conference on
		Integration of Constraint Programming, Artificial Intelligence, and
		Operations Research ({CPAIOR} 2018)}, pages 1--17, Delft, The Netherlands,
	2018. Springer.
	
	\bibitem{angluin87}
	D.~Angluin.
	\newblock Queries and concept learning.
	\newblock {\em Machine Learning}, 2(4):319--342, 1987.
	
	\bibitem{AngluinFP92}
	D.~Angluin, M.~Frazier, and L.~Pitt.
	\newblock Learning conjunctions of horn clauses.
	\newblock {\em Mach. Learn.}, 9:147--164, 1992.
	
	\bibitem{beletalTR05}
	N.~Beldiceanu, M.~Carlsson, and J.X. Rampon.
	\newblock Global constraint catalog.
	\newblock Technical Report T2005:08, Swedish Institute of Computer Science,
	Kista, Sweden, May 2005.
	
	\bibitem{DBLP:conf/cp/BeldiceanuS12}
	N.~Beldiceanu and H.~Simonis.
	\newblock A model seeker: Extracting global constraint models from positive
	examples.
	\newblock In {\em Proceedings of the Seventeenth International Conference on
		Principles and Practice of Constraint Programming (CP'12)}, pages 141--157,
	Quebec City, Canada, 2012. Springer.
	
	\bibitem{BessiereCDLMB14}
	C.~Bessiere, R.~Coletta, A.~Daoudi, N.~Lazaar, Y.~Mechqrane, and E.H. Bouyakhf.
	\newblock Boosting constraint acquisition via generalization queries.
	\newblock In {\em Proceedings of the 21st European Conference on Artificial
		Intelligence}, pages 99--104, Prague, Czech Republic, 2014. {IOS} Press.
	
	\bibitem{besetalCP04modelling}
	C.~Bessiere, R.~Coletta, E.~Freuder, and B.~O'Sullivan.
	\newblock Leveraging the learning power of examples in automated constraint
	acquisition.
	\newblock In {\em Proceedings of the Tenth International Conference on
		Principles and Practice of Constraint Programming (CP'04)}, pages 123--137,
	Toronto, Canada, 2004. Springer.
	
	\bibitem{BessiereCHKLNQW13}
	C.~Bessiere, R.~Coletta, E.~Hebrard, G.~Katsirelos, N.~Lazaar, N.~Narodytska,
	C.-G. Quimper, and T.~Walsh.
	\newblock Constraint acquisition via partial queries.
	\newblock In {\em Proceedings of the 23rd International Joint Conference on
		Artificial Intelligence}, pages 475--481, Beijing, China, 2013.
	
	\bibitem{besetalECML05}
	C.~Bessiere, R.~Coletta, F.~Koriche, and B.~O'Sullivan.
	\newblock A {SAT}-based version space algorithm for acquiring constraint
	satisfaction problems.
	\newblock In {\em Proceedings of the European Conference on Machine Learning
		(ECML'05)}, pages 23--34, Porto, Portugal, 2005. Springer.
	
	\bibitem{besetalIJCAI07queries}
	C.~Bessiere, R.~Coletta, B~O'Sullivan, and M.~Paulin.
	\newblock Query-driven constraint acquisition.
	\newblock In {\em Proceedings of the Twentieth International Joint Conference
		on Artificial Intelligence (IJCAI'07)}, pages 44--49, Hyderabad, India, 2007.
	
	\bibitem{beskor12}
	C.~Bessiere and F.~Koriche.
	\newblock Non learnability of constraint networks with membership queries.
	\newblock Technical report, Coconut, Montpellier, France, February 2012.
	
	\bibitem{BessiereKLO17}
	C.~Bessiere, F.~Koriche, N.~Lazaar, and B.~O'Sullivan.
	\newblock Constraint acquisition.
	\newblock {\em Artif. Intell.}, 244:315--342, 2017.
	
	\bibitem{bouetalECAI04}
	F.~Boussemart, F.~Hemery, C.~Lecoutre, and L.~Sais.
	\newblock Boosting systematic search by weighting constraints.
	\newblock In {\em Proceedings of the Sixteenth European Conference on
		Artificial Intelligence (ECAI'04)}, pages 146--150, Valencia, Spain, 2004.
	
	\bibitem{CabonGLSW99}
	B.~Cabon, S.~de~Givry, L.~Lobjois, T.~Schiex, and J.P. Warners.
	\newblock Radio link frequency assignment.
	\newblock {\em Constraints}, 4(1):79--89, 1999.
	
	\bibitem{daoetalIJCAI16}
	A.~Daoudi, Y.~Mechqrane, C.~Bessiere, N.~Lazaar, and E.H. Bouyakhf.
	\newblock Constraint acquisition using recommendation queries.
	\newblock In {\em Proceedings of the Twenty-Fifth International Joint
		Conference on Artificial Intelligence (IJCAI'16)}, pages 720--726, New York
	City, NY, 2016.
	
	\bibitem{de1970asymptotic}
	N.G. De~Bruijn.
	\newblock {\em {Asymptotic Methods in Analysis}}.
	\newblock Dover Books on Mathematics. Dover Publications, 1970.
	
	\bibitem{frewalCP98}
	E.C. Freuder and R.J. Wallace.
	\newblock Suggestion strategies for constraint-based matchmaker agents.
	\newblock In Michael~J. Maher and Jean{-}Francois Puget, editors, {\em
		Proceedings of the 4th International Conference on Principles and Practice of
		Constraint Programming (CP98)}, pages 192--204, Pisa, Italy, 1998. Springer.
	
	\bibitem{csplib}
	I.P. Gent and T.~Walsh.
	\newblock Csplib: a benchmark library for constraints.
	\newblock http://www.csplib.org/, 1999.
	
	\bibitem{junkerAAAI04}
	U.~Junker.
	\newblock Quickxplain: Preferred explanations and relaxations for
	over-constrained problems.
	\newblock In {\em Proceedings of the Nineteenth National Conference on
		Artificial Intelligence (AAAI'04)}, pages 167--172, San Jose CA, 2004.
	
	\bibitem{KumarTR19}
	M.~Kumar, S.~Teso, and L.~De Raedt.
	\newblock Acquiring integer programs from data.
	\newblock In {\em Proceedings of the Twenty-Eighth International Joint
		Conference on Artificial Intelligence ({IJCAI} 2019)}, pages 1130--1136,
	Macao, China, 2019. ijcai.org.
	
	\bibitem{DBLP:conf/ictai/LallouetLMV10}
	A.~Lallouet, M.~Lopez, L.~Martin, and C.~Vrain.
	\newblock On learning constraint problems.
	\newblock In {\em Proceedings of the 22nd IEEE International Conference on
		Tools for Artificial Intelligence (IEEE-ICTAI'10)}, pages 45--52, Arras,
	France, 2010.
	
	\bibitem{dell-magazine}
	J.~Mason.
	\newblock Purdey's general store.
	\newblock {\em Dell Magazine}, 54:10--10, April 1997.
	
	\bibitem{pauetalICTAI08}
	M.~Paulin, C.~Bessiere, and J.~Sallantin.
	\newblock Automatic design of robot behaviors through constraint network
	acquisition.
	\newblock In {\em Proceedings of the 20th IEEE International Conference on
		Tools for Artificial Intelligence (IEEE-ICTAI'08)}, pages 275--282, Dayton
	OH, 2008.
	
	\bibitem{RaedtPT18}
	L.~De Raedt, A.~Passerini, and S.~Teso.
	\newblock Learning constraints from examples.
	\newblock In {\em Proceedings of the Thirty-Second {AAAI} Conference on
		Artificial Intelligence (AAAI-18)}, pages 7965--7970, New Orleans, Louisiana,
	USA, 2018. {AAAI} Press.
	
	\bibitem{TsourosSB19}
	D.C. Tsouros, K.~Stergiou, and C.~Bessiere.
	\newblock Structure-driven multiple constraint acquisition.
	\newblock In {\em Proceedings of the 25th International Conference on
		Principles and Practice of Constraint Programming (CP 2019)}, pages 709--725,
	Stamford, CT, 2019. Springer.
	
	\bibitem{TsourosSS18}
	D.C. Tsouros, K.~Stergiou, and P.G. Sarigiannidis.
	\newblock Efficient methods for constraint acquisition.
	\newblock In John~N. Hooker, editor, {\em Proceedings of the 24th International
		Conference Principles and Practice of Constraint Programming ({CP} 2018)},
	pages 373--388, Lille, France, 2018. Springer.
	
\end{thebibliography}

\end{document}